\documentclass{article} 
\usepackage{iclr2024_conference,times}
\iclrfinalcopy

\usepackage[utf8]{inputenc} 

\usepackage[T1]{fontenc}    
\usepackage{hyperref}       
\usepackage{url}            
\usepackage{booktabs}       
\usepackage{amsfonts}       
\usepackage{nicefrac}       
\usepackage{microtype}      
\usepackage{xcolor}         
\usepackage{wrapfig}
\usepackage{graphicx}
\usepackage{amsmath,amsfonts,amsthm}       
\usepackage{mathtools}      
\usepackage{multirow}
\usepackage{diagbox}
\usepackage{array}
\PassOptionsToPackage{hyphens}{url}
\usepackage{hyperref}
\usepackage[capitalise]{cleveref}
\usepackage{comment}
\usepackage{etoolbox}
\usepackage{graphicx}
\usepackage{afterpage}
\usepackage{placeins}

\usepackage[font=small]{caption}
\usepackage{algorithm} 
\usepackage{algpseudocode}  

\usepackage{tabularx} 
\usepackage{authblk} 

\usepackage{tikz}
\usepackage{scalerel}
\usetikzlibrary{patterns}
\usepackage{bm}

\usepackage{graphicx}
\usepackage{booktabs} 
\usepackage{multirow} 
\usepackage{wrapfig}
\usepackage{subcaption}
\usepackage{svg}
\usepackage{ulem}

\usepackage{xcolor}
\usepackage{colortbl}
\usepackage{pifont}
\newcommand{\xmark}{\ding{55}}
\newcommand{\cmark}{\ding{51}}
\usepackage{makecell}

\title{HDLCoRe: A Training-Free Framework for Mitigating Hallucinations in LLM-Generated HDL}
\author[1*]{\textbf{Heng Ping}}
\author[1*]{\textbf{Shixuan Li}}
\author[1*]{\textbf{Peiyu Zhang}}
\author[1*]{\textbf{Anzhe Cheng}}
\author[1*]{\textbf{Shukai Duan}}
\author[1]{\textbf{Nikos Kanakaris}}
\author[1]{\textbf{Xiongye Xiao}}
\author[1]{\textbf{Wei Yang}}
\author[1]{\textbf{Shahin Nazarian}}
\author[1]{\textbf{Andrei Irimia}}
\author[1†]{\textbf{Paul Bogdan}}

\affil[1]{\centering University of Southern California, Los Angeles, CA 90089, USA}

\begin{document}
\maketitle

\begingroup
\renewcommand\thefootnote{}
\footnotetext{\textsuperscript{*}Equal Contribution. \textsuperscript{†}Corresponding to: \texttt{pbogdan@usc.edu}.}
\endgroup

\begin{abstract}

Recent advances in large language models (LLMs) have demonstrated remarkable capabilities in code generation tasks. However, when applied to hardware description languages (HDL), these models exhibit significant limitations due to data scarcity, resulting in hallucinations and incorrect code generation. To address these challenges, we propose HDLCoRe, a training-free framework that enhances LLMs' HDL generation capabilities through prompt engineering techniques and retrieval-augmented generation (RAG). Our approach consists of two main components: (1) an HDL-aware Chain-of-Thought (CoT) prompting technique with self-verification that classifies tasks by complexity and type, incorporates domain-specific knowledge, and guides LLMs through step-by-step self-simulation for error correction; and (2) a two-stage heterogeneous RAG system that addresses formatting inconsistencies through key component extraction and efficiently retrieves relevant HDL examples through sequential filtering and re-ranking. HDLCoRe eliminates the need for model fine-tuning while substantially improving LLMs' HDL generation capabilities. Experimental results demonstrate that our framework achieves superior performance on the RTLLM2.0 benchmark, significantly reducing hallucinations and improving both syntactic and functional correctness.
 
\end{abstract}

\allowdisplaybreaks


\section{Introduction}

With the rapid advancement of semiconductor technology, the design of very large-scale integration (VLSI) has become increasingly vital across industries~\cite{huang2021machine}. Hardware description language (HDL) code, as the foundation of VLSI design, plays a critical role in defining the circuit architecture and functionality~\cite{palnitkar2003verilog}. In recent years, large language models (LLMs) have experienced explosive growth and demonstrated extraordinary capabilities in many aspects~\cite{kanakaris2025networkinformedpromptengineeringorganized, li2025climatellmefficientweatherforecasting}, especially in automated code generation~\cite{brown2020language, chen2021evaluating}. LLMs like GPT-4~\cite{achiam2023gpt} and Claude~\cite{anthropic2024claude} not only match expert programmers in general-purpose languages such as Python and C++, but also show unexpected competence in HDL. Automating HDL code generation efficiently and accurately represents a significant research opportunity that could substantially accelerate hardware development cycles~\cite{sun2025paradigm, zhao2024codev}.

Despite their impressive performance, commercial closed-source LLMs raise important privacy concerns, especially in hardware design where proprietary HDL code often represents valuable intellectual property requiring protection~\cite{gohil2024llmpirate, liu2024openllm, liu2024rtlcoder2}. Open-source LLMs like Llama~\cite{touvron2023llama} and Qwen~\cite{bai2023qwen} offer a promising alternative that better safeguards user data. However, these open LLMs face significant challenges when generating HDL code. They show a stronger tendency towards hallucination compared to their performance with general-purpose languages~\cite{yang2025haven}. This means they often generate HDL code that appears plausible but contains syntax and functional errors. The primary reason is the limited HDL-specific data in their training corpora, leading to insufficient knowledge about hardware design principles~\cite{sun2024classification}.

Researchers have explored various approaches to enhance LLMs' HDL code generation capabilities~\cite{pinckney2025revisiting}. Fine-tuning methods construct specialized HDL datasets to adapt open-source LLMs, sometimes achieving or surpassing commercial LLMs in HDL code generation~\cite{thakur2023benchmarking, thakur2024verigen, liu2023chipnemo, pei2024betterv, liu2024rtlcoder, chang2024data, gao2024autovcoder, zhang2024mg}. However, these approaches typically demand substantial computational resources and complex training procedures, limiting their practicality for individual users or smaller teams. Consequently, training-free methods have gained attention~\cite{fu2023gpt4aigchip, huang2024towards, thakur2023autochip, nakkab2024rome, vijayaraghavan2024chain, zhao2025vrank}. Some leverage Chain-of-Thought (CoT)~\cite{wei2022chain} prompting to decompose the generation process into sequential reasoning steps~\cite{vijayaraghavan2024chain, lu2024rtllm}. Others introduce hardware paradigm knowledge, distinguishing between combinational and sequential logic~\cite{sun2025paradigm, sun2024classification}. Some incorporate simulator feedback through tools like VCS or Design Compiler~\cite{huang2024towards, thakur2023autochip, nakkab2024rome}. However, these methods either introduce only basic HDL knowledge or require multiple non-end-to-end iterations with external tools, complicating the development process.

To overcome these limitations, we propose HDLCoRe (HDL-aware CoT and Retrieval), an efficient training-free and tool-free framework that enhances open-source LLMs' HDL code generation capabilities through two primary components. First, an HDL-aware CoT prompting technique with self-verification classifies HDL tasks by complexity and type, then augments prompts with tailored HDL knowledge to guide initial code generation. This approach subsequently prompts the LLM to create corresponding testbenches and perform step-by-step self-simulation, enabling identification of functional issues and optimizations for code refinement—all within a structured CoT paradigm with intermediate reasoning steps. Second, our efficient two-stage RAG system establishes a comprehensive heterogeneous database from curated open-source HDL datasets and addresses formatting inconsistencies by extracting multiple key components from task descriptions. The system employs a novel retrieval process that first conducts broad filtering to identify potential candidates, then performs refined re-ranking to select the most relevant HDL code examples. This systematic approach provides contextually appropriate guidance to the LLM, enhancing its ability to generate accurate HDL implementations while maintaining computational efficiency throughout the generation process.

\textbf{Our contributions.} The key contributions of this work are as follows:
\begin{itemize}

    \item \textbf{HDL-CoT}: We introduce a novel HDL-aware CoT prompting technique with self-verification that systematically incorporates domain expertise. Our prompting technique integrates HDL knowledge specific to task classifications into prompts and implements an automatic testbench generation mechanism that enables internal verification and refinement of generated code.
    
    \item \textbf{Efficient Heterogeneous RAG}: We establish a comprehensive heterogeneous database from carefully selected open-source HDL datasets. Our proposed RAG system extracts multiple key components from task descriptions to address formatting inconsistencies across diverse sources, then employs a two-stage retrieval process that efficiently identifies optimal HDL examples through sequential filtering and re-ranking, providing contextually relevant guidance for target tasks.
    
    \item \textbf{Superior Performance}: Through extensive evaluation on the RTLLM2.0 benchmark, we demonstrate our framework significantly outperforms existing training-free and tool-free methods in both syntactic and functional correctness. Our framework shows substantial improvements over state-of-the-art approaches, achieving up to 28\% higher functional pass@1 rates.
\end{itemize}


\section{RELATED WORK}

\subsection{Fine-tuning Approaches for HDL Code Generation}

\noindent Despite privacy advantages, open-source LLMs struggle with hallucination in HDL code generation due to domain-specific data scarcity. Several fine-tuning approaches address this limitation through different dataset construction strategies~\cite{thakur2023benchmarking, thakur2024verigen, liu2023chipnemo, pei2024betterv, liu2024rtlcoder, chang2024data, gao2024autovcoder, zhang2024mg}. VerGen~\cite{thakur2024verigen}, Chipnemo~\cite{liu2023chipnemo} created large-scale HDL datasets by collecting and filtering publicly available code from GitHub repositories. Subsequently, instruction-based fine-tuning methods emerged to enhance task-specific performance. Among these, RTLCoder~\cite{liu2024rtlcoder} generated task descriptions and HDL implementations by prompting GPT-3.5 with hardware keywords, then filtering results through syntax checking. CodeV~\cite{zhao2024codev} reversed this approach by first collecting high-quality HDL code, then using GPT-3.5 to generate matching task descriptions. AutoVCoder~\cite{gao2024autovcoder} advanced the field by combining both synthetic and GitHub-extracted pairs in a two-round fine-tuning process, further enhanced by integrating an HDL-specific retrieval-augmented generation system. 

These fine-tuning methods have demonstrated significant improvements in reducing hallucination and enhancing HDL code quality. However, they typically require substantial computational resources and complex hyperparameter optimization, creating accessibility barriers for individual users and small teams with limited computing infrastructure.



\subsection{Training-free Methods for HDL Code Generation}
To overcome resource constraints, researchers developed training-free methods for enhancing HDL code generation~\cite{fu2023gpt4aigchip, huang2024towards, thakur2023autochip, nakkab2024rome, vijayaraghavan2024chain, zhao2025vrank}. Chain-of-Thought approaches like SP~\cite{lu2024rtllm}, CoDes~\cite{vijayaraghavan2024chain} decompose generation into sequential reasoning steps, while Paradigm-Based~\cite{sun2025paradigm} incorporates knowledge about combinational and sequential logic circuits in prompts. These methods guide LLMs through structured reasoning processes that better align with hardware design methodologies. Simulator-feedback methods offer another direction: Rome~\cite{nakkab2024rome} implements hierarchical prompting with progressive optimization based on simulator feedback; VeriAssist~\cite{huang2024towards} generates both code and testbenches for self-correction; and PromptV~\cite{mi2024promptv}, MAGE~\cite{zhao2024mage} employ specialized agents for collaborative HDL generation and debugging.

While these approaches eliminate training requirements, they either provide minimal domain knowledge or require multiple non-end-to-end iterations with external tools. Our work addresses these limitations through an HDL-aware CoT prompting technique with self-verification and an efficient two-stage heterogeneous RAG system that maintains end-to-end efficiency while incorporating rich domain knowledge.


\section{PROPOSED FRAMEWORK}

\begin{figure*}[!t]
    \centering
    \includegraphics[width=\textwidth]{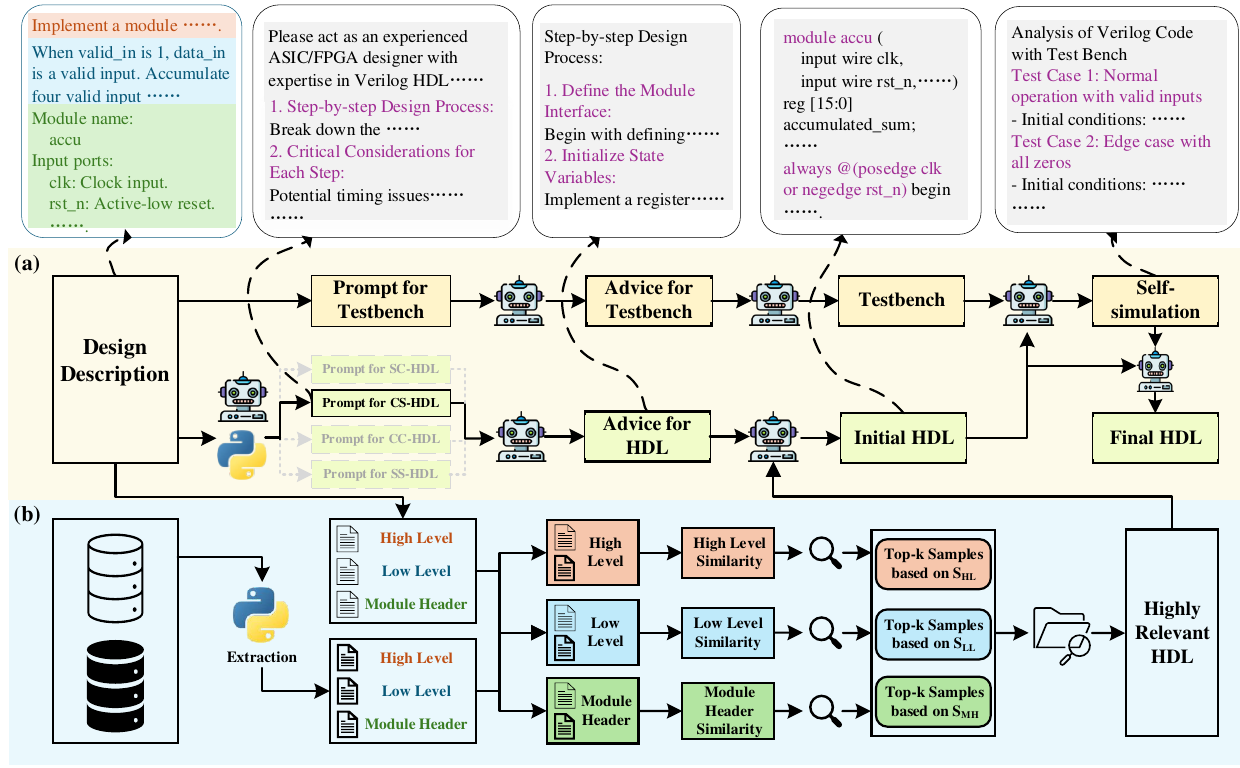}
    \caption{Overview of \textbf{HDLCoRe } framework. \textbf{(a) HDL-aware CoT with Self-verification Module.} We first process the design description through \textbf{task-specific prompts (SC-HDL, CC-HDL, SS-HDL, CS-HDL)} to generate initial HDL code. The framework then prompts for testbench creation followed by step-by-step self-simulation to identify errors and optimize the code. \textbf{(b) Efficient Heterogeneous RAG System.} We extract \textbf{multiple key components (High Level, Low Level, Module Header)} from both the task description and heterogeneous database, compute similarity scores ($S_{HL}$, $S_{LL}$, $S_{MH}$), and perform two-stage retrieval that first conducts broad filtering to select top-k samples from each category, then performs refined re-ranking to identify the most relevant HDL examples for the target task.}
    \label{fig:framework}
    \vspace{-4mm}
\end{figure*}

\noindent Figure~\ref{fig:framework} illustrates our proposed training-free and tool-free framework for enhancing LLMs' HDL code generation capabilities. This framework consists of two main components. The first component is a prompt technique based on an HDL-aware CoT mechanism, supplemented with self-verification to enhance error correction capabilities. The second is an efficient two-stage RAG system constructed from our heterogeneous HDL database. This system effectively identifies highly relevant HDL code examples from multi-source heterogeneous database and incorporates them into the generation context, thus augmenting the model's HDL generation proficiency. 

\subsection{HDL-Aware CoT Prompting with Self-verification}
\noindent As shown in Figure \ref{fig:framework}a, the HDL-aware CoT with self-verification mechanism consists of two primary parts. The first part involves classifying HDL code generation tasks and applying corresponding category-specific prompts to guide the LLM in generating initial HDL code based on the CoT mechanism. The second part prompts the LLM to create corresponding testbenches based on the task description. It then guides the LLM to perform a step-by-step simulation of the initial HDL code within these testbenches, enabling optimization and refinement of the original code based on self-simulation results.


\subsubsection{Specific CoT based on Task Type and Complexity}
While CoT mechanisms can enhance LLMs' code generation capabilities, their performance in HDL code generation remains limited due to the relative scarcity of HDL-related training corpora. Therefore, introducing domain-specific knowledge into the CoT process becomes essential. Inspired by~\cite{sun2025paradigm}, we categorize targeted HDL code generation tasks and incorporate knowledge specific to each category in the prompts, making the CoT process HDL-aware.

Specifically, we classify targeted HDL code generation tasks into four categories based on two criteria: simplicity versus complexity, and combinational versus sequential logic. These categories include Simple Combinational HDL (SC-HDL), Simple Sequential HDL (SS-HDL), Complex Combinational HDL (CC-HDL), and Complex Sequential HDL (CS-HDL). The distinction between combinational and sequential logic is addressed as follows: While the combinational logic primarily involves logical operations between signals that change simultaneously, the sequential logic involves temporal relationships between signals, requiring careful consideration of timing effects. To implement this classification, we use Python scripts to detect temporal signals in task descriptions, determining whether a task involves combinational or sequential logic.

Our preliminary experiments revealed that task complexity significantly impacts the prompt strategy. Simple HDL tasks perform better with minimal HDL-related knowledge in prompts, as excessive constraints often introduce errors. In contrast, complex tasks benefit from comprehensive HDL-related knowledge to prevent misunderstandings and functional omissions. For complexity assessment, we leverage the LLM itself to determine whether additional knowledge is required based on its capabilities. This dual-criteria classification enables us to provide appropriate guidance based on both logic type and task difficulty, significantly improving code quality across varying HDL implementation requirements.

\begin{figure*}[!t]
    \centering
    \includegraphics[width=\textwidth]{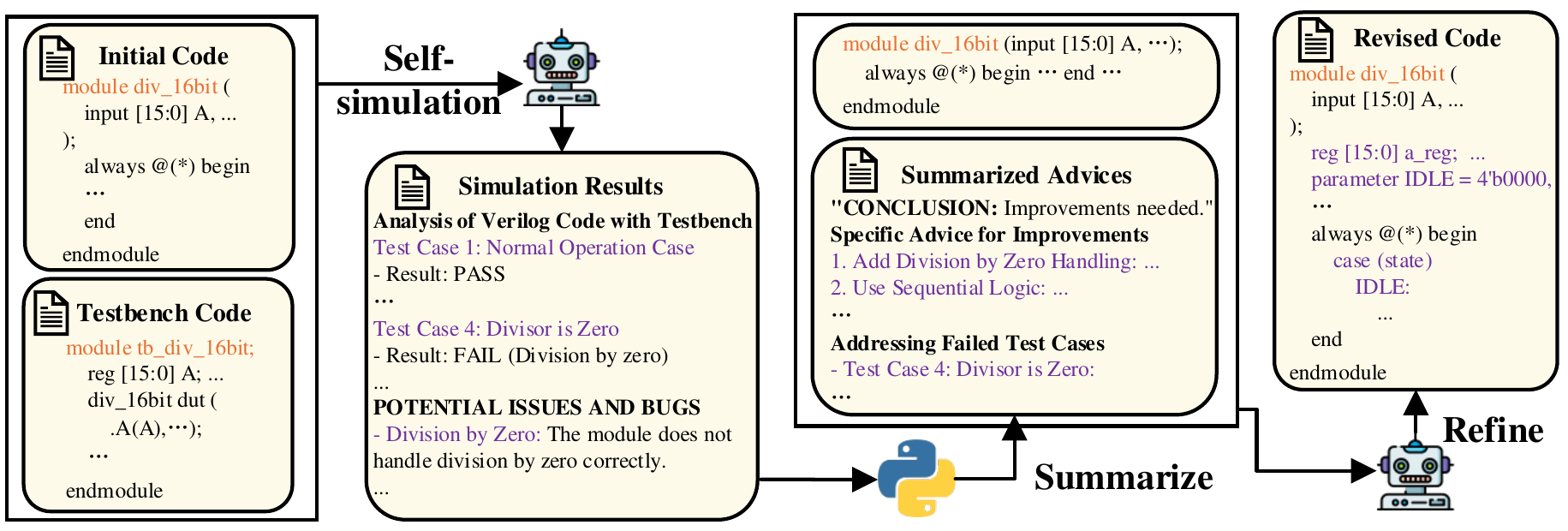}
    \caption{\textbf{Self-verification mechanism}. The LLM generates initial HDL code and testbench, performs \textbf{step-by-step self-simulation}, \textbf{summarizes the results}, and \textbf{refines the code based on identified issues}.}
    \label{fig:self-verification}
\end{figure*}

\subsubsection{Self-verification through step-by-step self-simulation}

Prior research~\cite{huang2024towards, thakur2023autochip, nakkab2024rome} has demonstrated that validating LLM-generated HDL code through simulators and feeding back the simulation results significantly enhances the LLM's ability to detect, correct and optimize HDL codes. Based on this insight, we incorporate a self-verification process within our HDL-aware CoT, enabling LLMs to generate testbenches autonomously and emulate actual simulators when evaluating their initially generated HDL code. The self-verification mechanism leverages the inherent reasoning capabilities of LLMs to examine and enhance the initially generated HDL code.

Figure \ref{fig:self-verification} illustrates our proposed self-verification approach. The process begins with the LLM generating both initial HDL code and corresponding testbench. These are then combined into a single prompt that instructs the LLM to validate the code against the testcases, documenting its verification process in a specified format. A Python script extracts and summarizes the detailed simulation output. This summary and the initial code are then fed back to the LLM to produce the final optimized implementation. This structured summarization technique mitigates information overload that would occur with complex simulation processes, enhancing the effectiveness of the self-verification while offering a more streamlined process compared to methods relying on actual simulators.

\subsection{Efficient Two-stage Heterogeneous RAG System}

\noindent Retrieval-augmented generation (RAG) operates by identifying relevant examples from an established database and presenting them to LLMs~\cite{borgeaud2022improving}. This approach enables the LLMs to learn domain-specific knowledge and enhance their performance on targeted tasks~\cite{wang2023survey}. Integrating RAG into prompt engineering can significantly improve LLMs' capabilities in addressing domain-specific challenges. The effectiveness of RAG primarily depends on two factors: the quality of the constructed database and the efficiency of the retrieval methodology~\cite{zhao2024towards}. Based on these considerations, we have developed an efficient two-stage heterogeneous RAG system specifically tailored for HDL code generation to reduce hallucinations and strengthen the LLMs' ability to generate accurate HDL codes.

In our system, we first establish a comprehensive heterogeneous database by collecting and filtering HDL code generation-related resources from open-source repositories. This curated collection serves as a high-quality knowledge base for the RAG system. To address the challenges of retrieving information from multi-source heterogeneous databases, we designed a two-stage retrieval algorithm based on multiple key component matching. This approach effectively mitigates the impact of formatting inconsistencies between data from different sources while efficiently retrieving HDL code examples highly relevant to the target task. Our algorithm prioritizes semantic relevance over superficial similarities, ensuring that retrieved examples provide meaningful guidance to the LLM regardless of their original formatting or structure.

\subsubsection{Construction of Comprehensive Heterogeneous Database}
Previous research~\cite{thakur2023benchmarking, thakur2024verigen, liu2023chipnemo, pei2024betterv, liu2024rtlcoder, chang2024data, gao2024autovcoder, zhang2024mg} has explored various fine-tuning approaches to enhance LLMs' HDL code generation capabilities. These works typically construct dedicated datasets to serve as training data for fine-tuning LLMs. Such datasets generally follow an instruction-code pattern, where each data contains an instruction component describing the task and an HDL code component providing the solution. This structured format aligns well with our target RAG system, which operates by computing similarity between the target task and instruction components to retrieve appropriate HDL code examples.

The quality of these HDL-related datasets varies significantly, with some generated by ChatGPT and others manually crafted by human experts. Dataset size and diversity also differ considerably across sources. To address these variations, we conducted thorough examination and filtering, ultimately selecting five high-quality open-source datasets~\cite{liu2024rtlcoder, zhang2024mg, 2024origen, yang2025haven, gao2024autovcoder} to compose our comprehensive database. This curated collection forms a robust foundation highly relevant to HDL code generation tasks.

\subsubsection{Efficient Two-stage Retrieval Method for Heterogeneous Database}
As illustrated in Figure \ref{fig:framework}b, we propose a two-stage retrieval method to efficiently extract HDL code examples highly relevant to target tasks from heterogeneous databases. Our approach addresses formatting inconsistencies across different data sources through multiple key component matching. The method begins with preprocessing instruction text to extract three key components highly relevant to HDL code generation: the high-level task overview, low-level implementation details, and the module header containing input/output signals. These critical components are extracted using Python scripts and provide a basis for semantic matching that mitigates formatting differences across heterogeneous data sources.

Our two-stage retrieval process balances efficiency and quality through coarse-grained filtering followed by fine-grained selection. In the first stage, we employ a specialized text embedding model (e.g., gtr-t5-large) to semantically embed the three components, obtaining high-level embeddings $H_{HL}$, low-level embeddings $H_{LL}$, and module header embeddings $H_{MH}$. We calculate cosine similarity scores ($S_{HL}$, $S_{LL}$, and $S_{MH}$) between the target task and database samples, selecting the top-k most relevant samples from each category to form an initial filtered database of 3k samples.

In the second stage, we use a cross-encoder (e.g., bge-reranker-v2-m3) to perform detailed re-ranking of these filtered samples. The cross-encoder computes cross-attention between the complete instruction content of each filtered sample and the target task, producing refined similarity scores that capture complex semantic interactions. This allows us to select the top-N most relevant examples as the final exemplars provided to the LLM, enhancing the identification of associations between HDL function descriptions and signal definitions. This sequential approach effectively combines the efficiency of lightweight embedding models for initial filtering with the precision of complex cross-encoders for final selection, guaranteeing both speed and quality in retrieval performance.

\section{EXPERIMENTS}

\noindent In this section, we conduct a comprehensive evaluation of our framework through a series of experiments designed to address three principal research questions (RQs):

\textbf{RQ1}: How does our framework's HDL code generation capability compare with state-of-the-art (SOTA) training-free and tool-free methods?

\textbf{RQ2}: How does our framework perform across LLMs of varying scales and categories (chat models vs. code models)?

\textbf{RQ3}: What contributions do the core components of our framework make to its overall effectiveness?

\subsection{Experimental Setup}
\noindent We evaluate our framework on RTLLM 2.0~\cite{liu2024openllm}, a comprehensive benchmark consisting of 50 Verilog design tasks spanning various complexity levels. We compare our approach against SOTA training-free and tool-free methods: Self-Planning (SP)~\cite{lu2024rtllm} and Chain-of-Descriptions (CoDes)~\cite{vijayaraghavan2024chain}. To investigate performance variations across LLM scales and categories, we employ general-purpose LLMs (Qwen2.5 series in 7B, 14B, 32B, and 72B sizes) and code-specialized LLMs (Qwen2.5-Coder series in 7B, 14B, and 32B sizes). These models enable us to evaluate how LLM size and domain specialization affect HDL code generation capabilities.

We assess both syntax correctness (Syn.) and functional correctness (Fun.) using Synopsys VCS, an industry-standard HDL simulator and verification tool. Specifically, we use VCS for syntax checking through compilation and functional verification through simulation testing. Only HDL code that passes syntax checking proceeds to functional verification. We evaluate using two metrics: \textbf{pass@1}, which measures success rate with a single sampling attempt, simulating real-world deployment scenarios; and \textbf{pass@5}, which measures success rate when the best result from five sampling attempts is considered. For pass@1 evaluation, we set a very low temperature value to ensure deterministic outputs and minimize randomness, while for pass@5, we use a moderate temperature setting to allow for creative variations. Following~\cite{lu2024rtllm}, a task is considered successful if any generation attempt passes the verification.

\begin{table*}[!t]
\centering
\setlength{\tabcolsep}{4pt} 
\large 
\caption{Performance Comparison of Different Training-free and Tool-free Methods on RTLLM2.0 Dataset for HDL Code Generation. \textcolor{orange}{Orange} and \textcolor{cyan}{Cyan} Colors indicate the \textcolor{orange}{Best} and \textcolor{cyan}{Second-best} Results, Respectively.}
\resizebox{\textwidth}{!}{ 
\begin{tabular}{l|c|c|c|c|c|c|c|c|c|c|c|c|c|c|c|c}
\Xhline{1pt}
\multirow{3}{*}{Model} & \multicolumn{4}{c|}{Direct (\%)} & \multicolumn{4}{c|}{SP (\%)} & \multicolumn{4}{c|}{CoDes (\%)} & \multicolumn{4}{c}{HDLCoRe (\%)} \\
\cline{2-17}
& \multicolumn{2}{c|}{pass@1} & \multicolumn{2}{c|}{pass@5} & \multicolumn{2}{c|}{pass@1} & \multicolumn{2}{c|}{pass@5} & \multicolumn{2}{c|}{pass@1} & \multicolumn{2}{c|}{pass@5} & \multicolumn{2}{c|}{pass@1} & \multicolumn{2}{c}{pass@5} \\
\cline{2-17}
& Syn. & Fun. & Syn. & Fun. & Syn. & Fun. & Syn. & Fun. & Syn. & Fun. & Syn. & Fun. & Syn. & Fun. & Syn. & Fun. \\
\Xhline{0.8pt}
qwen2.5:7b & \textcolor{cyan}{62} & 12 & \textcolor{cyan}{68} & 18 & 50 & \textcolor{cyan}{20} & 62 & \textcolor{cyan}{26} & 46 & 14 & 60 & 22 & \textcolor{orange}{64} & \textcolor{orange}{40} & \textcolor{orange}{70} & \textcolor{orange}{44} \\
qwen2.5-coder:7b & \textcolor{cyan}{74} & 24 & \textcolor{orange}{78} & 26 & 66 & \textcolor{cyan}{30} & 68 & \textcolor{cyan}{32} & 54 & 28 & 66 & 32 & \textcolor{orange}{74} & \textcolor{orange}{40} & \textcolor{cyan}{76} & \textcolor{orange}{46} \\
qwen2.5:14b & \textcolor{orange}{78} & 26 & \textcolor{orange}{80} & 30 & 62 & \textcolor{cyan}{36} & 64 & \textcolor{cyan}{38} & \textcolor{cyan}{74} & 30 & 74 & 36 & 68 & \textcolor{orange}{42} & \textcolor{cyan}{76} & \textcolor{orange}{46} \\
qwen2.5-coder:14b & \textcolor{orange}{84} & 36 & \textcolor{orange}{88} & 42 & \textcolor{cyan}{76} & \textcolor{cyan}{44} & 80 & 48 & \textcolor{cyan}{76} & 42 & 82 & \textcolor{cyan}{48} & 74 & \textcolor{orange}{48} & \textcolor{cyan}{82} & \textcolor{orange}{54} \\
qwen2.5:32b & 78 & 42 & \textcolor{cyan}{86} & 46 & \textcolor{cyan}{82} & \textcolor{cyan}{52} & 84 & \textcolor{cyan}{56} & 76 & 46 & 80 & 52 & \textcolor{orange}{86} & \textcolor{orange}{60} & \textcolor{orange}{88} & \textcolor{orange}{62} \\
qwen2.5-coder:32b & 84 & 40 & \textcolor{cyan}{86} & 46 & \textcolor{cyan}{84} & \textcolor{cyan}{50} & 84 & \textcolor{cyan}{52} & 62 & 44 & 76 & 50 & \textcolor{orange}{84} & \textcolor{orange}{54} & \textcolor{orange}{86} & \textcolor{orange}{60} \\
qwen2.5:72b & \textcolor{orange}{86} & 42 & \textcolor{orange}{88} & 44 & 78 & \textcolor{cyan}{46} & 82 & \textcolor{cyan}{48} & 70 & 44 & 80 & 48 & \textcolor{cyan}{78} & \textcolor{orange}{52} & \textcolor{cyan}{84} & \textcolor{orange}{56} \\
\Xhline{1pt}
\end{tabular}
}

\label{tab:main_result}
\end{table*}

\subsection{Results for HDL Code Generation (RQ1)}
\noindent Table~\ref{tab:main_result} presents a performance comparison between our framework and baseline methods for HDL code generation, where the Direct approach represents the baseline method with HDL code generation task descriptions input directly into LLMs without additional processing. Our results demonstrate that across all tested LLMs, regardless of size or category, our framework achieves favorable syntax pass rates and superior functional pass rates in both pass@1 and pass@5 tests, with functional pass rates showing substantial improvements compared to the baselines.

A detailed examination of our experimental results validates the effectiveness of our framework and reveals several important findings that support our design choices. First, the Direct baseline often exhibits high syntax pass rates but low functional pass rates. This discrepancy occurs because when task descriptions are directly used as prompts without additional guidance, LLMs tend to generate structurally simple HDL code. Such code contains fewer syntax errors due to its simplicity but frequently fails to implement the required functionality correctly.

Among the baselines, SP generally outperforms CoDes. This performance difference can be attributed to SP's self-planning approach, which allows LLMs to determine HDL code generation strategies on their own. In contrast, CoDes requires LLMs to follow more complex patterns when providing HDL code generation guidance. For relatively smaller LLMs (ranging from 7B to 72B parameters), SP's approach is less likely to cause information overload, thereby reducing hallucinations during HDL code generation. This observation validates our framework's classification-based CoT technique to HDL code generation tasks, which enables LLMs to assess task difficulty relative to their own capabilities.

When comparing our framework to the SP and CoDes baselines, we observe significant improvements in functional pass rates. These improvements stem primarily from two key components of our approach: the efficient heterogeneous RAG system and the self-verification technique. The RAG system provides precise example HDL code to guide the LLMs, while the self-verification technique optimizes the initial generation results. Together, these components effectively reduce hallucinations in LLM-generated HDL code and lower error rates.

\subsection{Performance across Scales and Categories (RQ2)}
\noindent Table~\ref{tab:main_result} reveals key insights about our framework's performance across different LLMs. Larger LLMs consistently outperform smaller ones, with \emph{qwen2.5:32b} achieving 86\% syntax and 60\% functional pass@1 rates (88\% and 62\% for pass@5) compared to 64\% and 40\% for \emph{qwen2.5:7b} (70\% and 44\% for pass@5). Code-specialized LLMs generally demonstrate better HDL generation capabilities than general-purpose LLMs of comparable size, particularly for syntax correctness.

Notably, our framework provides greater relative improvements for smaller LLMs compared to larger ones. Smaller LLMs receive more substantial benefits from our approach: \emph{qwen2.5:7b} shows a 28\% absolute improvement in functional pass@1 rate over the Direct baseline (26\% for pass@5), while \emph{qwen2.5:72b} exhibits a 10\% improvement (12\% for pass@5). This stems from our classification-based CoT preventing the information overload in smaller LLMs and our RAG system compensating for their limited HDL prior knowledge. Additionally, our framework effectively narrows the performance gap between general and code-specialized LLMs, suggesting that our framework successfully compensates for the lack of code-specific pre-training.

\begin{figure*}[!t]
    \centering
    \includegraphics[width=\textwidth]{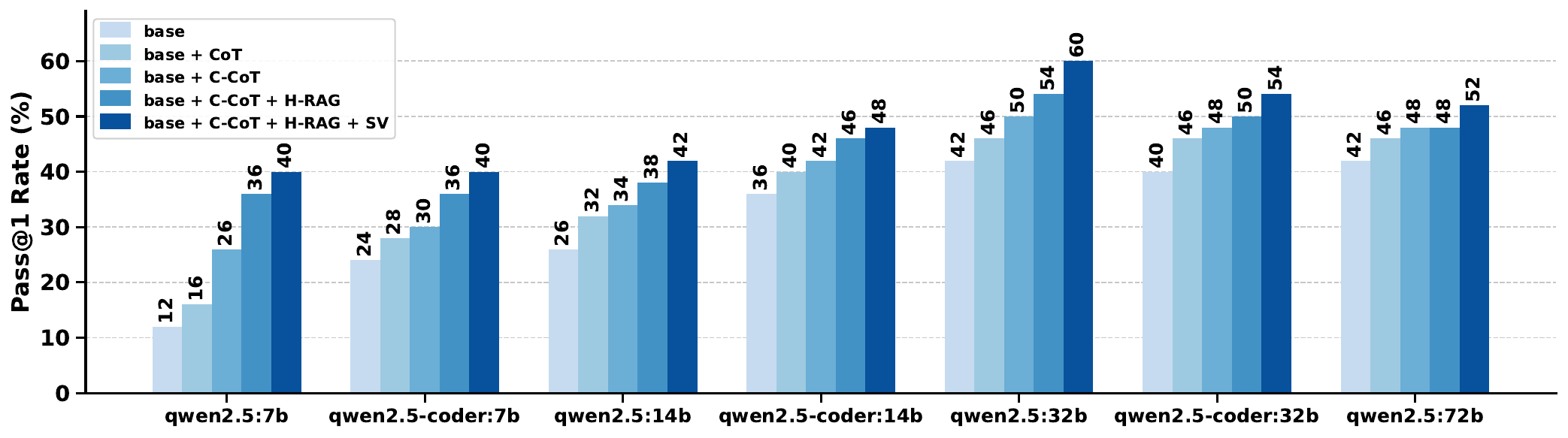}
    \caption{Ablation Study of techniques adopted in our framework. Pass@1 performance of functional correctness is reported on RTLLM2.0 dataset.}
    \label{fig:ablation}
\vspace{-3mm}
\end{figure*}

\subsection{Ablation Study (RQ3)}
\noindent We perform comparative analysis with pass@1 functional correctness as the evaluation metric across five configurations:
\begin{itemize}
    \item \textbf{Base:} LLMs without any additional techniques;
    \item \textbf{Base+CoT:} Extending LLMs with simple CoT ;
    \item \textbf{Base+C-CoT:} Incorporating classification-based CoT;
    \item \textbf{Base+C-CoT+H-RAG:} Integrating classification-based CoT with our efficient heterogeneous RAG system;
    \item \textbf{Base+C-CoT+H-RAG+SV:} The comprehensive framework including self-verification mechanisms.
\end{itemize}

Figure \ref{fig:ablation} illustrates the contribution of each core component in our framework to enhancing LLMs' HDL code generation capabilities. Our classification-based CoT demonstrates a 2-10\% improvement over traditional CoT approaches, confirming the effectiveness of tailoring reasoning processes to specific HDL tasks. The efficient heterogeneous RAG system provides up to 10\% improvement for smaller LLMs (\emph{qwen2.5:7b}), highlighting the importance of providing precise HDL code examples to LLMs with limited size. Additionally, self-verification consistently improves performance across all tested LLMs by 2-6\%, demonstrating that having models verify their own code correctness is universally beneficial. Overall, the ablation study validates the effectiveness of each component, with our complete framework achieving the best performance across all model variants.


\section{Conclusion and Future Work}
In this paper, we present an efficient training-free and tool-free framework for enhancing the HDL code generation capabilities of open-source LLMs. Our approach combines an HDL-aware CoT prompting technique with self-verification and an efficient two-stage heterogeneous RAG system to provide contextually relevant guidance. Extensive evaluation on the RTLLM2.0 benchmark demonstrates that our framework significantly outperforms state-of-the-art methods, with particularly substantial improvements for smaller LLMs. The ablation study confirms that each key component contributes meaningfully to overall performance. Our work establishes a computationally efficient path for enhancing HDL code generation without fine-tuning resources or external tools, making advanced hardware design more accessible to individual users and smaller teams. Future work could explore extending this approach to develop techniques for automatically generating HDL code optimized for performance, power, and area (PPA) constraints.


\clearpage


\medskip

\clearpage
{
\small
\bibliography{iclr2024_conference}

\begin{thebibliography}{40}
\providecommand{\natexlab}[1]{#1}
\providecommand{\url}[1]{\texttt{#1}}
\expandafter\ifx\csname urlstyle\endcsname\relax
  \providecommand{\doi}[1]{doi: #1}\else
  \providecommand{\doi}{doi: \begingroup \urlstyle{rm}\Url}\fi

\bibitem[Achiam et~al.(2023)Achiam, Adler, Agarwal, Ahmad, Akkaya, Aleman, Almeida, Altenschmidt, Altman, Anadkat, Avila, et~al.]{achiam2023gpt}
Josh Achiam, Steven Adler, Sandhini Agarwal, Lama Ahmad, Ilge Akkaya, Florencia~Leoni Aleman, Diogo Almeida, Johannes Altenschmidt, Sam Altman, Sonal Anadkat, Rene Avila, et~al.
\newblock Gpt-4 technical report.
\newblock \emph{arXiv preprint arXiv:2303.08774}, 2023.

\bibitem[Anthropic(2024)]{anthropic2024claude}
Anthropic.
\newblock The claude 3 model family: Opus, sonnet, haiku.
\newblock In \emph{Technical Report}, 2024.
\newblock URL \url{https://api.semanticscholar.org/CorpusID:268232499}.

\bibitem[Bai et~al.(2023)Bai, Bai, Chu, Cui, Dang, Deng, Fan, Ge, Han, Huang, Hui, et~al.]{bai2023qwen}
Jinze Bai, Shuai Bai, Yunfei Chu, Zeyu Cui, Kai Dang, Xiaodong Deng, Yang Fan, Wei Ge, Yu~Han, Fei Huang, Binyuan Hui, et~al.
\newblock Qwen technical report.
\newblock \emph{arXiv preprint arXiv:2309.16609}, 2023.

\bibitem[Borgeaud et~al.(2022)Borgeaud, Mensch, Hoffmann, Cai, Rutherford, Millican, Van Den~Driessche, Lespiau, Damoc, Clark, de~Las~Casas, et~al.]{borgeaud2022improving}
Sebastian Borgeaud, Arthur Mensch, Jordan Hoffmann, Trevor Cai, Eliza Rutherford, Katie Millican, George~Bm Van Den~Driessche, Jean-Baptiste Lespiau, Bogdan Damoc, Aidan Clark, Diego de~Las~Casas, et~al.
\newblock Improving language models by retrieving from trillions of tokens.
\newblock In \emph{International conference on machine learning}, pp.\  2206--2240. PMLR, 2022.

\bibitem[Brown et~al.(2020)Brown, Mann, Ryder, Subbiah, Kaplan, Dhariwal, Neelakantan, Shyam, Sastry, Askell, Agarwal, and Amodei]{brown2020language}
Tom Brown, Benjamin Mann, Nick Ryder, Melanie Subbiah, Jared~D Kaplan, Prafulla Dhariwal, Arvind Neelakantan, Pranav Shyam, Girish Sastry, Amanda Askell, Sandhini Agarwal, and Dario Amodei.
\newblock Language models are few-shot learners.
\newblock \emph{Advances in neural information processing systems}, 33:\penalty0 1877--1901, 2020.

\bibitem[Chang et~al.(2024)Chang, Wang, Yang, Wang, Jin, Zhu, Chen, Li, Yan, Zhou, Zhao, and Li]{chang2024data}
Kaiyan Chang, Kun Wang, Nan Yang, Ying Wang, Dantong Jin, Wenlong Zhu, Zhirong Chen, Chong Li, Han Yan, Yuan Zhou, Zijian Zhao, and Xiao Li.
\newblock Data is all you need: Finetuning llms for chip design via an automated design-data augmentation framework.
\newblock In \emph{Proceedings of the 61st ACM/IEEE Design Automation Conference}, pp.\  1--6. ACM/IEEE, 2024.

\bibitem[Chen et~al.(2021)Chen, Tworek, Jun, Yuan, Pinto, Kaplan, Edwards, Burda, Joseph, Brockman, Ray, and Zaremba]{chen2021evaluating}
Mark Chen, Jerry Tworek, Heewoo Jun, Qiming Yuan, Henrique Ponde De~Oliveira Pinto, Jared Kaplan, Harri Edwards, Yura Burda, Nicholas Joseph, Greg Brockman, Alex Ray, and Wojciech Zaremba.
\newblock Evaluating large language models trained on code.
\newblock \emph{arXiv preprint arXiv:2107.03374}, 2021.

\bibitem[Cui et~al.(2024)Cui, Yin, Zhou, Xiao, Sun, Xu, Guo, Song, Lin, Zhang, et~al.]{2024origen}
Fan Cui, Chenyang Yin, Kexing Zhou, Youwei Xiao, Guangyu Sun, Qiang Xu, Qipeng Guo, Demin Song, Dahua Lin, Xingcheng Zhang, et~al.
\newblock Origen: Enhancing rtl code generation with code-to-code augmentation and self-reflection.
\newblock \emph{arXiv preprint arXiv:2407.16237}, 2024.

\bibitem[Fu et~al.(2023)Fu, Zhang, Yu, Li, Ye, Li, Wan, and Lin]{fu2023gpt4aigchip}
Yonggan Fu, Yongan Zhang, Zhongzhi Yu, Sixu Li, Zhifan Ye, Chaojian Li, Cheng Wan, and Yingyan~Celine Lin.
\newblock Gpt4aigchip: Towards next-generation ai accelerator design automation via large language models.
\newblock In \emph{2023 IEEE/ACM International Conference on Computer Aided Design (ICCAD)}, pp.\  1--9. IEEE, 2023.

\bibitem[Gao et~al.(2024)Gao, Zhao, Lin, Ding, Hou, Feng, Li, and Guo]{gao2024autovcoder}
Mingzhe Gao, Jieru Zhao, Zhe Lin, Wenchao Ding, Xiaofeng Hou, Yu~Feng, Chao Li, and Minyi Guo.
\newblock Autovcoder: A systematic framework for automated verilog code generation using llms.
\newblock In \emph{2024 IEEE 42nd International Conference on Computer Design (ICCD)}, pp.\  162--169. IEEE, 2024.

\bibitem[Gohil et~al.(2024)Gohil, DeLorenzo, Nallam, See, and Rajendran]{gohil2024llmpirate}
Vasudev Gohil, Matthew DeLorenzo, Veera Vishwa Achuta Sai~Venkat Nallam, Joey See, and Jeyavijayan Rajendran.
\newblock Llmpirate: Llms for black-box hardware ip piracy.
\newblock \emph{arXiv preprint arXiv:2411.16111}, 2024.

\bibitem[Huang et~al.(2021)Huang, Hu, He, Liu, Ma, Shen, Wu, Xu, Zhang, Zhong, Ning, and Wang]{huang2021machine}
Guyue Huang, Jingbo Hu, Yifan He, Jialong Liu, Mingyuan Ma, Zhaoyang Shen, Juejian Wu, Yunhui Xu, Hongce Zhang, Kai Zhong, Xin Ning, and Yu~Wang.
\newblock Machine learning for electronic design automation: A survey.
\newblock \emph{ACM Transactions on Design Automation of Electronic Systems (TODAES)}, 26\penalty0 (5):\penalty0 1--46, 2021.

\bibitem[Huang et~al.(2024)Huang, Lin, Wang, Chen, Ding, and Zhao]{huang2024towards}
Hanxian Huang, Zhenghan Lin, Zixuan Wang, Xin Chen, Ke~Ding, and Jishen Zhao.
\newblock Towards llm-powered verilog rtl assistant: Self-verification and self-correction.
\newblock \emph{arXiv preprint arXiv:2406.00115}, 2024.

\bibitem[Kanakaris et~al.(2025)Kanakaris, Ping, Xiao, Ahmed, Luceri, Ferrara, and Bogdan]{kanakaris2025networkinformedpromptengineeringorganized}
Nikos Kanakaris, Heng Ping, Xiongye Xiao, Nesreen~K. Ahmed, Luca Luceri, Emilio Ferrara, and Paul Bogdan.
\newblock Network-informed prompt engineering against organized astroturf campaigns under extreme class imbalance, 2025.
\newblock URL \url{https://arxiv.org/abs/2501.11849}.

\bibitem[Li et~al.(2025)Li, Yang, Zhang, Xiao, Cao, Qin, Zhang, Zhao, and Bogdan]{li2025climatellmefficientweatherforecasting}
Shixuan Li, Wei Yang, Peiyu Zhang, Xiongye Xiao, Defu Cao, Yuehan Qin, Xiaole Zhang, Yue Zhao, and Paul Bogdan.
\newblock Climatellm: Efficient weather forecasting via frequency-aware large language models, 2025.
\newblock URL \url{https://arxiv.org/abs/2502.11059}.

\bibitem[Liu et~al.(2023)Liu, Ene, Kirby, Cheng, Pinckney, Liang, Alben, Anand, Banerjee, Bayraktaroglu, Bhaskaran, et~al.]{liu2023chipnemo}
Mingjie Liu, Teodor-Dumitru Ene, Robert Kirby, Chris Cheng, Nathaniel Pinckney, Rongjian Liang, Jonah Alben, Hari Anand, Supratik Banerjee, Ilknur Bayraktaroglu, Bharath Bhaskaran, et~al.
\newblock Chipnemo: Domain-adapted llms for chip design.
\newblock \emph{arXiv preprint arXiv:2311.00176}, 2023.

\bibitem[Liu et~al.(2024{\natexlab{a}})Liu, Fang, Lu, Wang, Zhang, Zhang, and Xie]{liu2024rtlcoder2}
Shang Liu, Wenji Fang, Yao Lu, Jing Wang, Qijun Zhang, Hongce Zhang, and Zhiyao Xie.
\newblock Rtlcoder: Fully open-source and efficient llm-assisted rtl code generation technique.
\newblock \emph{IEEE Transactions on Computer-Aided Design of Integrated Circuits and Systems}, 2024{\natexlab{a}}.

\bibitem[Liu et~al.(2024{\natexlab{b}})Liu, Fang, Lu, Zhang, Zhang, and Xie]{liu2024rtlcoder}
Shang Liu, Wenji Fang, Yao Lu, Qijun Zhang, Hongce Zhang, and Zhiyao Xie.
\newblock Rtlcoder: Outperforming gpt-3.5 in design rtl generation with our open-source dataset and lightweight solution.
\newblock In \emph{2024 IEEE LLM Aided Design Workshop (LAD)}, pp.\  1--5. IEEE, 2024{\natexlab{b}}.

\bibitem[Liu et~al.(2024{\natexlab{c}})Liu, Lu, Fang, Li, and Xie]{liu2024openllm}
Shang Liu, Yao Lu, Wenji Fang, Mengming Li, and Zhiyao Xie.
\newblock Openllm-rtl: Open dataset and benchmark for llm-aided design rtl generation.
\newblock In \emph{IEEE/ACM International Conference on Computer-Aided Design (ICCAD)}. IEEE, 2024{\natexlab{c}}.

\bibitem[Lu et~al.(2024)Lu, Liu, Zhang, and Xie]{lu2024rtllm}
Yao Lu, Shang Liu, Qijun Zhang, and Zhiyao Xie.
\newblock Rtllm: An open-source benchmark for design rtl generation with large language model.
\newblock In \emph{2024 29th Asia and South Pacific Design Automation Conference (ASP-DAC)}, pp.\  722--727. IEEE, 2024.

\bibitem[Mi et~al.(2024)Mi, Zheng, Zhong, Sun, and Huang]{mi2024promptv}
Zhendong Mi, Renming Zheng, Haowen Zhong, Yue Sun, and Shaoyi Huang.
\newblock Promptv: Leveraging llm-powered multi-agent prompting for high-quality verilog generation.
\newblock \emph{arXiv preprint arXiv:2412.11014}, 2024.

\bibitem[Nakkab et~al.(2024)Nakkab, Zhang, Karri, and Garg]{nakkab2024rome}
Andre Nakkab, Sai~Qian Zhang, Ramesh Karri, and Siddharth Garg.
\newblock Rome was not built in a single step: Hierarchical prompting for llm-based chip design.
\newblock In \emph{Proceedings of the 2024 ACM/IEEE International Symposium on Machine Learning for CAD}, pp.\  1--11. ACM/IEEE, 2024.

\bibitem[Palnitkar(2003)]{palnitkar2003verilog}
Samir Palnitkar.
\newblock \emph{Verilog HDL: a guide to digital design and synthesis}, volume~1.
\newblock Prentice Hall Professional, 2003.

\bibitem[Pei et~al.(2024)Pei, Zhen, Yuan, Huang, and Yu]{pei2024betterv}
Zehua Pei, Hui-Ling Zhen, Mingxuan Yuan, Yu~Huang, and Bei Yu.
\newblock Betterv: Controlled verilog generation with discriminative guidance.
\newblock \emph{arXiv preprint arXiv:2402.03375}, 2024.

\bibitem[Pinckney et~al.(2025)Pinckney, Batten, Liu, Ren, and Khailany]{pinckney2025revisiting}
Nathaniel Pinckney, Christopher Batten, Mingjie Liu, Haoxing Ren, and Brucek Khailany.
\newblock Revisiting verilogeval: A year of improvements in large-language models for hardware code generation.
\newblock \emph{ACM Transactions on Design Automation of Electronic Systems}, 2025.

\bibitem[Sun et~al.(2024)Sun, Li, Zhang, Yin, Zhuo, and Schlichtmann]{sun2024classification}
Wenhao Sun, Bing Li, Grace~Li Zhang, Xunzhao Yin, Cheng Zhuo, and Ulf Schlichtmann.
\newblock Classification-based automatic hdl code generation using llms.
\newblock \emph{arXiv preprint arXiv:2407.18326}, 2024.

\bibitem[Sun et~al.(2025)Sun, Li, Zhang, Yin, Zhuo, and Schlichtmann]{sun2025paradigm}
Wenhao Sun, Bing Li, Grace~Li Zhang, Xunzhao Yin, Cheng Zhuo, and Ulf Schlichtmann.
\newblock Paradigm-based automatic hdl code generation using llms.
\newblock \emph{arXiv preprint arXiv:2501.12702}, 2025.

\bibitem[Thakur et~al.(2023{\natexlab{a}})Thakur, Ahmad, Fan, Pearce, Tan, Karri, Dolan-Gavitt, and Garg]{thakur2023benchmarking}
Shailja Thakur, Baleegh Ahmad, Zhenxing Fan, Hammond Pearce, Benjamin Tan, Ramesh Karri, Brendan Dolan-Gavitt, and Siddharth Garg.
\newblock Benchmarking large language models for automated verilog rtl code generation.
\newblock In \emph{2023 Design, Automation \& Test in Europe Conference \& Exhibition (DATE)}, pp.\  1--6. IEEE, 2023{\natexlab{a}}.

\bibitem[Thakur et~al.(2023{\natexlab{b}})Thakur, Blocklove, Pearce, Tan, Garg, and Karri]{thakur2023autochip}
Shailja Thakur, Jason Blocklove, Hammond Pearce, Benjamin Tan, Siddharth Garg, and Ramesh Karri.
\newblock Autochip: Automating hdl generation using llm feedback.
\newblock \emph{arXiv preprint arXiv:2311.04887}, 2023{\natexlab{b}}.

\bibitem[Thakur et~al.(2024)Thakur, Ahmad, Pearce, Tan, Dolan-Gavitt, Karri, and Garg]{thakur2024verigen}
Shailja Thakur, Baleegh Ahmad, Hammond Pearce, Benjamin Tan, Brendan Dolan-Gavitt, Ramesh Karri, and Siddharth Garg.
\newblock Verigen: A large language model for verilog code generation.
\newblock \emph{ACM Transactions on Design Automation of Electronic Systems}, 29\penalty0 (3):\penalty0 1--31, 2024.

\bibitem[Touvron et~al.(2023)Touvron, Lavril, Izacard, Martinet, Lachaux, Lacroix, Rozi{\`e}re, Goyal, Hambro, Azhar, Rodriguez, et~al.]{touvron2023llama}
Hugo Touvron, Thibaut Lavril, Gautier Izacard, Xavier Martinet, Marie-Anne Lachaux, Timoth{\'e}e Lacroix, Baptiste Rozi{\`e}re, Naman Goyal, Eric Hambro, Faisal Azhar, Aurelien Rodriguez, et~al.
\newblock Llama: Open and efficient foundation language models.
\newblock \emph{arXiv preprint arXiv:2302.13971}, 2023.

\bibitem[Vijayaraghavan et~al.(2024)Vijayaraghavan, Nitsure, Mackin, Shi, Ambrogio, Haran, Paruthi, Elzein, Coops, Beymer, and Baldwin]{vijayaraghavan2024chain}
Prashanth Vijayaraghavan, Apoorva Nitsure, Charles Mackin, Luyao Shi, Stefano Ambrogio, Arvind Haran, Viresh Paruthi, Abdeladim Elzein, Derek Coops, David Beymer, and Timothee Baldwin.
\newblock Chain-of-descriptions: Improving code llms for vhdl code generation and summarization.
\newblock In \emph{Proceedings of the 2024 ACM/IEEE International Symposium on Machine Learning for CAD}, pp.\  1--10. ACM/IEEE, 2024.

\bibitem[Wang et~al.(2023)Wang, Liu, Yue, Tang, Zhang, Jiayang, Yao, Gao, Hu, Qi, Wang, et~al.]{wang2023survey}
Cunxiang Wang, Xiaoze Liu, Yuanhao Yue, Xiangru Tang, Tianhang Zhang, Cheng Jiayang, Yunzhi Yao, Wenqi Gao, Xuming Hu, Zengzhi Qi, Yue Wang, et~al.
\newblock Survey on factuality in large language models: Knowledge, retrieval and domain-specificity.
\newblock \emph{arXiv preprint arXiv:2310.07521}, 2023.

\bibitem[Wei et~al.(2022)Wei, Wang, Schuurmans, Bosma, Xia, Chi, Le, and Zhou]{wei2022chain}
Jason Wei, Xuezhi Wang, Dale Schuurmans, Maarten Bosma, Fei Xia, Ed~Chi, Quoc~V Le, and Denny Zhou.
\newblock Chain-of-thought prompting elicits reasoning in large language models.
\newblock \emph{Advances in neural information processing systems}, 35:\penalty0 24824--24837, 2022.

\bibitem[Yang et~al.(2025)Yang, Teng, Liu, Qi, Lv, Li, Zhang, and He]{yang2025haven}
Yiyao Yang, Fu~Teng, Pengju Liu, Mengnan Qi, Chenyang Lv, Ji~Li, Xuhong Zhang, and Zhezhi He.
\newblock Haven: Hallucination-mitigated llm for verilog code generation aligned with hdl engineers.
\newblock \emph{arXiv preprint arXiv:2501.04908}, 2025.

\bibitem[Zhang et~al.(2024)Zhang, Yu, Fu, Wan, and Lin]{zhang2024mg}
Yongan Zhang, Zhongzhi Yu, Yonggan Fu, Cheng Wan, and Yingyan~Celine Lin.
\newblock Mg-verilog: Multi-grained dataset towards enhanced llm-assisted verilog generation.
\newblock In \emph{2024 IEEE LLM Aided Design Workshop (LAD)}, pp.\  1--5. IEEE, 2024.

\bibitem[Zhao et~al.(2024{\natexlab{a}})Zhao, Huang, Song, Wang, Wan, and Ma]{zhao2024towards}
Shengming Zhao, Yuheng Huang, Jiayang Song, Zhijie Wang, Chengcheng Wan, and Lei Ma.
\newblock Towards understanding retrieval accuracy and prompt quality in rag systems.
\newblock \emph{arXiv preprint arXiv:2411.19463}, 2024{\natexlab{a}}.

\bibitem[Zhao et~al.(2024{\natexlab{b}})Zhao, Huang, Li, Jin, Nan, Ma, Qi, Pan, Zhang, Zhang, Zhang, and Chen]{zhao2024codev}
Yang Zhao, Di~Huang, Chongxiao Li, Pengwei Jin, Ziyuan Nan, Tianyun Ma, Lei Qi, Yu~Pan, Zhiwei Zhang, Ruizhi Zhang, Xuran Zhang, and Yiran Chen.
\newblock Codev: Empowering llms for verilog generation through multi-level summarization.
\newblock \emph{arXiv preprint arXiv:2407.10424}, 2024{\natexlab{b}}.

\bibitem[Zhao et~al.(2024{\natexlab{c}})Zhao, Zhang, Huang, Yu, and Zhao]{zhao2024mage}
Yujie Zhao, Hejia Zhang, Hanxian Huang, Zhongming Yu, and Jishen Zhao.
\newblock Mage: A multi-agent engine for automated rtl code generation.
\newblock \emph{arXiv preprint arXiv:2412.07822}, 2024{\natexlab{c}}.

\bibitem[Zhao et~al.(2025)Zhao, Qiu, Lin, Zhang, Li, and Schlichtmann]{zhao2025vrank}
Zhuorui Zhao, Ruidi Qiu, Ing-Chao Lin, Grace~Li Zhang, Bing Li, and Ulf Schlichtmann.
\newblock Vrank: Enhancing verilog code generation from large language models via self-consistency.
\newblock \emph{arXiv preprint arXiv:2502.00028}, 2025.

\end{thebibliography}
\bibliographystyle{iclr2024_conference}
}

\clearpage

\appendix

\section{Details of Benchmark RTLLM2.0}

RTLLM 2.0 is a comprehensive open-source benchmark designed to evaluate the capability of large language models (LLMs) in generating Hardware Description Language (HDL) code from natural language specifications. It is an extension of the original RTLLM benchmark, expanding the design collection from 30 to 50 diverse RTL designs.

Each design in the RTLLM 2.0 benchmark is accompanied by three essential files that facilitate comprehensive evaluation:
\begin{enumerate}
    \item \textbf{Design Description (design\_description.txt):} A natural language description of the target design's functionality, serving as a prompt for LLMs. It includes the module name and all input/output signal specifications with signal names and widths.
    
    \item \textbf{Testbench (testbench.v):} A Verilog testbench containing multiple test cases with input and expected output values, designed to verify the functional correctness of generated RTL code.
    
    \item \textbf{Reference Design (designer\_RTL.v):} A correct Verilog implementation hand-crafted by human designers, providing a golden reference against which the quality of automatically generated designs can be measured.
\end{enumerate}

The RTLLM 2.0 benchmark evaluates RTL generation capabilities according to three progressive goals, with \textbf{syntax correctness} and \textbf{functional correctness} being primary focus areas for our framework:
\begin{itemize}
    \item \textbf{Syntax Goal:} The syntax of the generated RTL design must be correct, verified by successful synthesis into a netlist without errors. \textit{This is a fundamental requirement and a primary evaluation metric for our framework, as syntactically incorrect designs cannot be implemented in hardware.}
    
    \item \textbf{Functionality Goal:} The generated RTL design must function as intended, verified by passing all test cases in the provided testbench. \textit{Our framework particularly focuses on improving this aspect, as functional correctness ensures the generated design actually implements the specified behavior.}
    
    \item \textbf{Quality Goal:} The generated RTL design should exhibit desirable design qualities in terms of performance, power, and area (PPA), measured after synthesis and layout.
\end{itemize}

In our work, we concentrate specifically on improving the syntax and functionality aspects of LLM-generated HDL code, as these are prerequisite conditions for any practical hardware implementation. The benchmark allows for a systematic, quantitative evaluation of LLM-based RTL generation techniques across designs of varying complexity. By providing both the natural language specifications and reference implementations, RTLLM 2.0 facilitates fair comparisons between different LLM solutions and enables the development of more advanced techniques for automatic hardware design.

Table~\ref{tab:rtllm_benchmark} presents the comprehensive design collection in RTLLM 2.0, categorized into four functional domains. For each design, we quantify its complexity using two key metrics:

\begin{table}[!t]
\caption{RTLLM 2.0 Benchmark Description and Complexity Metrics}
\label{tab:rtllm_benchmark}
\centering
\small
\setlength{\tabcolsep}{4pt}
\begin{tabular}{|l|p{4.5cm}|c|c|}
\hline
\textbf{Category} & \textbf{Design} & \textbf{LoC} & \textbf{Cells} \\
\hline
\multirow{11}{*}{Arithmetic} 
& adder\_8bit (An 8-bit adder) & 26 & 58 \\
& adder\_16bit (16-bit with full adders) & 137 & 130 \\
& adder\_32bit (32-bit carry-lookahead) & 181 & 312 \\
& adder\_pipe\_64bit (64-bit pipeline adder) & 197 & 1340 \\
& adder\_bcd (BCD adder for decimal arithmetic) & 64 & 195 \\
& sub\_64bit (64-bit subtractor) & 128 & 121 \\
& multi\_8bit (8-bit booth-4 multiplier) & 84 & 34 \\
& multi\_16bit (16-bit multiplier) & 65 & 817 \\
& multi\_booth\_8bit (8-bit booth multiplier) & 72 & 94 \\
& multi\_pipe\_4bit (4-bit pipeline multiplier) & 43 & 120 \\
& multi\_pipe\_8bit (8-bit pipeline multiplier) & 92 & 578 \\
\hline
\multirow{8}{*}{Memory} 
& asyn\_fifo (Asynchronous FIFO 16×8 bits) & 149 & 686 \\
& LIFObuffer (Last-In-First-Out buffer) & 52 & 135 \\
& right\_shifter (Right shifter with 8-bit delay) & 17 & 466 \\
& LFSR (Linear Feedback Shift Register) & 37 & 38 \\
& barrel\_shifter (Barrel shifter) & 46 & 19 \\
& RAM (8x4 bits true dual-port RAM) & 50 & 1834 \\
& ROM (Read-Only Memory) & 37 & 121 \\
\hline
\multirow{6}{*}{Control} 
& fsm (FSM detection circuit) & 77 & 24 \\
& sequence\_detector (Sequence detector) & 38 & 6 \\
& counter\_12 (12-bit counter) & 37 & 38 \\
& JC\_counter (4-bit Johnson counter) & 22 & 134 \\
& ring\_counter (8-bit ring counter) & 39 & 7 \\
& up\_down\_counter (16-bit up/down counter) & 111 & 2435 \\
\hline
\multirow{10}{*}{Misc.} 
& clkgenerator (Clock generator) & 49 & 64 \\
& instr\_reg (Instruction register) & 106 & 117 \\
& signal\_generator (Signal generator) & 52 & 135 \\
& square\_wave (Square wave generator) & 50 & 117 \\
& alu (ALU for 32bit MIPS-ISA CPU) & 111 & 2435 \\
& pe (Multiplying Accumulator) & 27 & 1439 \\
& freq\_div (Frequency divider) & 51 & 64 \\
& calendar (Perpetual calendar) & 37 & 121 \\
& traffic\_light (Traffic light system) & 106 & 117 \\
& width\_8to16 (8-bit to 16-bit converter) & 50 & 117 \\
\hline
\end{tabular}
\end{table}

\begin{table*}[!t]
\caption{Pass@1 Performance of HDLCoRe on RTLLM 2.0 Designs: Syntax and Functional Verification Results across Qwen2.5 LLMs}
\label{tab:verification_results_nonc}
\centering
\begin{tabular}{|l|cc|cc|cc|cc|}
\hline
\multirow{2}{*}{\textbf{Design}} & \multicolumn{2}{c|}{\textbf{Qwen2.5:7b}} & \multicolumn{2}{c|}{\textbf{Qwen2.5:14b}} & \multicolumn{2}{c|}{\textbf{Qwen2.5:32b}} & \multicolumn{2}{c|}{\textbf{Qwen2.5:72b}} \\
\cline{2-9}
 & Syn. & Fun. & Syn. & Fun. & Syn. & Fun. & Syn. & Fun. \\
\hline
                   accu & \cellcolor{green!25}\cmark &   \cellcolor{red!25}\xmark & \cellcolor{green!25}\cmark &   \cellcolor{red!25}\xmark & \cellcolor{green!25}\cmark & \cellcolor{green!25}\cmark & \cellcolor{green!25}\cmark &   \cellcolor{red!25}\xmark \\
            adder\_8bit  & \cellcolor{green!25}\cmark & \cellcolor{green!25}\cmark & \cellcolor{green!25}\cmark & \cellcolor{green!25}\cmark & \cellcolor{green!25}\cmark & \cellcolor{green!25}\cmark & \cellcolor{green!25}\cmark & \cellcolor{green!25}\cmark \\
            adder\_16bit & \cellcolor{green!25}\cmark & \cellcolor{green!25}\cmark & \cellcolor{green!25}\cmark & \cellcolor{green!25}\cmark & \cellcolor{green!25}\cmark & \cellcolor{green!25}\cmark & \cellcolor{green!25}\cmark & \cellcolor{green!25}\cmark \\
            adder\_32bit &   \cellcolor{red!25}\xmark &   \cellcolor{red!25}\xmark &   \cellcolor{red!25}\xmark &   \cellcolor{red!25}\xmark & \cellcolor{green!25}\cmark & \cellcolor{green!25}\cmark & \cellcolor{green!25}\cmark & \cellcolor{green!25}\cmark \\
              adder\_bcd & \cellcolor{green!25}\cmark & \cellcolor{green!25}\cmark & \cellcolor{green!25}\cmark & \cellcolor{green!25}\cmark & \cellcolor{green!25}\cmark & \cellcolor{green!25}\cmark & \cellcolor{green!25}\cmark &   \cellcolor{red!25}\xmark \\
       adder\_pipe\_64bit &   \cellcolor{red!25}\xmark &   \cellcolor{red!25}\xmark & \cellcolor{green!25}\cmark &   \cellcolor{red!25}\xmark & \cellcolor{green!25}\cmark & \cellcolor{green!25}\cmark &   \cellcolor{red!25}\xmark &   \cellcolor{red!25}\xmark \\
        comparator\_3bit & \cellcolor{green!25}\cmark & \cellcolor{green!25}\cmark & \cellcolor{green!25}\cmark & \cellcolor{green!25}\cmark & \cellcolor{green!25}\cmark & \cellcolor{green!25}\cmark & \cellcolor{green!25}\cmark & \cellcolor{green!25}\cmark \\
        comparator\_4bit & \cellcolor{green!25}\cmark & \cellcolor{green!25}\cmark & \cellcolor{green!25}\cmark & \cellcolor{green!25}\cmark & \cellcolor{green!25}\cmark &   \cellcolor{red!25}\xmark & \cellcolor{green!25}\cmark & \cellcolor{green!25}\cmark \\
              div\_16bit &   \cellcolor{red!25}\xmark &   \cellcolor{red!25}\xmark & \cellcolor{green!25}\cmark &   \cellcolor{red!25}\xmark & \cellcolor{green!25}\cmark &   \cellcolor{red!25}\xmark &   \cellcolor{red!25}\xmark &   \cellcolor{red!25}\xmark \\
             radix2\_div &   \cellcolor{red!25}\xmark &   \cellcolor{red!25}\xmark &   \cellcolor{red!25}\xmark &   \cellcolor{red!25}\xmark &   \cellcolor{red!25}\xmark &   \cellcolor{red!25}\xmark &   \cellcolor{red!25}\xmark &   \cellcolor{red!25}\xmark \\
             multi\_8bit &   \cellcolor{red!25}\xmark &   \cellcolor{red!25}\xmark &   \cellcolor{red!25}\xmark &   \cellcolor{red!25}\xmark & \cellcolor{green!25}\cmark & \cellcolor{green!25}\cmark & \cellcolor{green!25}\cmark &   \cellcolor{red!25}\xmark \\
            multi\_16bit & \cellcolor{green!25}\cmark &   \cellcolor{red!25}\xmark & \cellcolor{green!25}\cmark &   \cellcolor{red!25}\xmark & \cellcolor{green!25}\cmark & \cellcolor{green!25}\cmark & \cellcolor{green!25}\cmark & \cellcolor{green!25}\cmark \\
       multi\_booth\_8bit &   \cellcolor{red!25}\xmark &   \cellcolor{red!25}\xmark &   \cellcolor{red!25}\xmark &   \cellcolor{red!25}\xmark & \cellcolor{green!25}\cmark &   \cellcolor{red!25}\xmark & \cellcolor{green!25}\cmark & \cellcolor{green!25}\cmark \\
        multi\_pipe\_4bit & \cellcolor{green!25}\cmark &   \cellcolor{red!25}\xmark &   \cellcolor{red!25}\xmark &   \cellcolor{red!25}\xmark & \cellcolor{green!25}\cmark &   \cellcolor{red!25}\xmark & \cellcolor{green!25}\cmark &   \cellcolor{red!25}\xmark \\
        multi\_pipe\_8bit &   \cellcolor{red!25}\xmark &   \cellcolor{red!25}\xmark &   \cellcolor{red!25}\xmark &   \cellcolor{red!25}\xmark &   \cellcolor{red!25}\xmark &   \cellcolor{red!25}\xmark & \cellcolor{green!25}\cmark &   \cellcolor{red!25}\xmark \\
      fixed\_point\_adder & \cellcolor{green!25}\cmark & \cellcolor{green!25}\cmark & \cellcolor{green!25}\cmark &   \cellcolor{red!25}\xmark & \cellcolor{green!25}\cmark & \cellcolor{green!25}\cmark & \cellcolor{green!25}\cmark & \cellcolor{green!25}\cmark \\
fixed\_point\_substractor & \cellcolor{green!25}\cmark & \cellcolor{green!25}\cmark &   \cellcolor{red!25}\xmark &   \cellcolor{red!25}\xmark & \cellcolor{green!25}\cmark & \cellcolor{green!25}\cmark & \cellcolor{green!25}\cmark & \cellcolor{green!25}\cmark \\
            float\_multi & \cellcolor{green!25}\cmark &   \cellcolor{red!25}\xmark &   \cellcolor{red!25}\xmark &   \cellcolor{red!25}\xmark & \cellcolor{green!25}\cmark &   \cellcolor{red!25}\xmark & \cellcolor{green!25}\cmark &   \cellcolor{red!25}\xmark \\
              sub\_64bit & \cellcolor{green!25}\cmark & \cellcolor{green!25}\cmark & \cellcolor{green!25}\cmark & \cellcolor{green!25}\cmark & \cellcolor{green!25}\cmark & \cellcolor{green!25}\cmark &   \cellcolor{red!25}\xmark &   \cellcolor{red!25}\xmark \\
\hline
             counter\_12 & \cellcolor{green!25}\cmark & \cellcolor{green!25}\cmark & \cellcolor{green!25}\cmark & \cellcolor{green!25}\cmark & \cellcolor{green!25}\cmark & \cellcolor{green!25}\cmark & \cellcolor{green!25}\cmark & \cellcolor{green!25}\cmark \\
             JC\_counter & \cellcolor{green!25}\cmark &   \cellcolor{red!25}\xmark & \cellcolor{green!25}\cmark & \cellcolor{green!25}\cmark & \cellcolor{green!25}\cmark & \cellcolor{green!25}\cmark & \cellcolor{green!25}\cmark & \cellcolor{green!25}\cmark \\
           ring\_counter & \cellcolor{green!25}\cmark & \cellcolor{green!25}\cmark & \cellcolor{green!25}\cmark & \cellcolor{green!25}\cmark & \cellcolor{green!25}\cmark & \cellcolor{green!25}\cmark & \cellcolor{green!25}\cmark & \cellcolor{green!25}\cmark \\
       up\_down\_counter  & \cellcolor{green!25}\cmark & \cellcolor{green!25}\cmark & \cellcolor{green!25}\cmark & \cellcolor{green!25}\cmark & \cellcolor{green!25}\cmark & \cellcolor{green!25}\cmark & \cellcolor{green!25}\cmark &   \cellcolor{red!25}\xmark \\
                    fsm &   \cellcolor{red!25}\xmark &   \cellcolor{red!25}\xmark & \cellcolor{green!25}\cmark &   \cellcolor{red!25}\xmark & \cellcolor{green!25}\cmark &   \cellcolor{red!25}\xmark &   \cellcolor{red!25}\xmark &   \cellcolor{red!25}\xmark \\
      sequence\_detector &   \cellcolor{red!25}\xmark &   \cellcolor{red!25}\xmark &   \cellcolor{red!25}\xmark &   \cellcolor{red!25}\xmark &   \cellcolor{red!25}\xmark &   \cellcolor{red!25}\xmark &   \cellcolor{red!25}\xmark &   \cellcolor{red!25}\xmark \\
          right\_shifter & \cellcolor{green!25}\cmark & \cellcolor{green!25}\cmark & \cellcolor{green!25}\cmark & \cellcolor{green!25}\cmark & \cellcolor{green!25}\cmark & \cellcolor{green!25}\cmark & \cellcolor{green!25}\cmark & \cellcolor{green!25}\cmark \\
               freq\_div & \cellcolor{green!25}\cmark &   \cellcolor{red!25}\xmark & \cellcolor{green!25}\cmark & \cellcolor{green!25}\cmark & \cellcolor{green!25}\cmark & \cellcolor{green!25}\cmark & \cellcolor{green!25}\cmark & \cellcolor{green!25}\cmark \\
         freq\_divbyeven &   \cellcolor{red!25}\xmark &   \cellcolor{red!25}\xmark &   \cellcolor{red!25}\xmark &   \cellcolor{red!25}\xmark &   \cellcolor{red!25}\xmark &   \cellcolor{red!25}\xmark &   \cellcolor{red!25}\xmark &   \cellcolor{red!25}\xmark \\
         freq\_divbyfrac &   \cellcolor{red!25}\xmark &   \cellcolor{red!25}\xmark & \cellcolor{green!25}\cmark &   \cellcolor{red!25}\xmark & \cellcolor{green!25}\cmark &   \cellcolor{red!25}\xmark & \cellcolor{green!25}\cmark &   \cellcolor{red!25}\xmark \\
          freq\_divbyodd &   \cellcolor{red!25}\xmark &   \cellcolor{red!25}\xmark &   \cellcolor{red!25}\xmark &   \cellcolor{red!25}\xmark & \cellcolor{green!25}\cmark &   \cellcolor{red!25}\xmark &   \cellcolor{red!25}\xmark &   \cellcolor{red!25}\xmark \\
\hline
              asyn\_fifo &   \cellcolor{red!25}\xmark &   \cellcolor{red!25}\xmark &   \cellcolor{red!25}\xmark &   \cellcolor{red!25}\xmark &   \cellcolor{red!25}\xmark &   \cellcolor{red!25}\xmark &   \cellcolor{red!25}\xmark &   \cellcolor{red!25}\xmark \\
             LIFObuffer &   \cellcolor{red!25}\xmark &   \cellcolor{red!25}\xmark & \cellcolor{green!25}\cmark &   \cellcolor{red!25}\xmark & \cellcolor{green!25}\cmark &   \cellcolor{red!25}\xmark & \cellcolor{green!25}\cmark &   \cellcolor{red!25}\xmark \\
         barrel\_shifter &   \cellcolor{red!25}\xmark &   \cellcolor{red!25}\xmark &   \cellcolor{red!25}\xmark &   \cellcolor{red!25}\xmark & \cellcolor{green!25}\cmark &   \cellcolor{red!25}\xmark &   \cellcolor{red!25}\xmark &   \cellcolor{red!25}\xmark \\
                   LFSR &   \cellcolor{red!25}\xmark &   \cellcolor{red!25}\xmark &   \cellcolor{red!25}\xmark &   \cellcolor{red!25}\xmark &   \cellcolor{red!25}\xmark &   \cellcolor{red!25}\xmark &   \cellcolor{red!25}\xmark &   \cellcolor{red!25}\xmark \\
                    RAM & \cellcolor{green!25}\cmark & \cellcolor{green!25}\cmark & \cellcolor{green!25}\cmark & \cellcolor{green!25}\cmark & \cellcolor{green!25}\cmark & \cellcolor{green!25}\cmark & \cellcolor{green!25}\cmark & \cellcolor{green!25}\cmark \\
                    ROM & \cellcolor{green!25}\cmark & \cellcolor{green!25}\cmark & \cellcolor{green!25}\cmark & \cellcolor{green!25}\cmark & \cellcolor{green!25}\cmark & \cellcolor{green!25}\cmark & \cellcolor{green!25}\cmark & \cellcolor{green!25}\cmark \\
\hline
               calendar & \cellcolor{green!25}\cmark & \cellcolor{green!25}\cmark & \cellcolor{green!25}\cmark & \cellcolor{green!25}\cmark & \cellcolor{green!25}\cmark & \cellcolor{green!25}\cmark & \cellcolor{green!25}\cmark & \cellcolor{green!25}\cmark \\
            edge\_detect & \cellcolor{green!25}\cmark &   \cellcolor{red!25}\xmark & \cellcolor{green!25}\cmark & \cellcolor{green!25}\cmark & \cellcolor{green!25}\cmark & \cellcolor{green!25}\cmark & \cellcolor{green!25}\cmark & \cellcolor{green!25}\cmark \\
        parallel2serial & \cellcolor{green!25}\cmark &   \cellcolor{red!25}\xmark & \cellcolor{green!25}\cmark &   \cellcolor{red!25}\xmark & \cellcolor{green!25}\cmark &   \cellcolor{red!25}\xmark & \cellcolor{green!25}\cmark &   \cellcolor{red!25}\xmark \\
           pulse\_detect & \cellcolor{green!25}\cmark &   \cellcolor{red!25}\xmark & \cellcolor{green!25}\cmark &   \cellcolor{red!25}\xmark & \cellcolor{green!25}\cmark &   \cellcolor{red!25}\xmark & \cellcolor{green!25}\cmark &   \cellcolor{red!25}\xmark \\
        serial2parallel & \cellcolor{green!25}\cmark &   \cellcolor{red!25}\xmark & \cellcolor{green!25}\cmark &   \cellcolor{red!25}\xmark & \cellcolor{green!25}\cmark & \cellcolor{green!25}\cmark & \cellcolor{green!25}\cmark & \cellcolor{green!25}\cmark \\
           synchronizer & \cellcolor{green!25}\cmark & \cellcolor{green!25}\cmark & \cellcolor{green!25}\cmark & \cellcolor{green!25}\cmark & \cellcolor{green!25}\cmark & \cellcolor{green!25}\cmark & \cellcolor{green!25}\cmark & \cellcolor{green!25}\cmark \\
          traffic\_light &   \cellcolor{red!25}\xmark &   \cellcolor{red!25}\xmark &   \cellcolor{red!25}\xmark &   \cellcolor{red!25}\xmark & \cellcolor{green!25}\cmark &   \cellcolor{red!25}\xmark & \cellcolor{green!25}\cmark &   \cellcolor{red!25}\xmark \\
            width\_8to16 & \cellcolor{green!25}\cmark &   \cellcolor{red!25}\xmark & \cellcolor{green!25}\cmark &   \cellcolor{red!25}\xmark & \cellcolor{green!25}\cmark & \cellcolor{green!25}\cmark & \cellcolor{green!25}\cmark & \cellcolor{green!25}\cmark \\
                    alu &   \cellcolor{red!25}\xmark &   \cellcolor{red!25}\xmark &   \cellcolor{red!25}\xmark &   \cellcolor{red!25}\xmark &   \cellcolor{red!25}\xmark &   \cellcolor{red!25}\xmark & \cellcolor{green!25}\cmark &   \cellcolor{red!25}\xmark \\
           clkgenerator & \cellcolor{green!25}\cmark & \cellcolor{green!25}\cmark & \cellcolor{green!25}\cmark & \cellcolor{green!25}\cmark & \cellcolor{green!25}\cmark & \cellcolor{green!25}\cmark & \cellcolor{green!25}\cmark & \cellcolor{green!25}\cmark \\
              instr\_reg & \cellcolor{green!25}\cmark &   \cellcolor{red!25}\xmark & \cellcolor{green!25}\cmark &   \cellcolor{red!25}\xmark & \cellcolor{green!25}\cmark & \cellcolor{green!25}\cmark & \cellcolor{green!25}\cmark & \cellcolor{green!25}\cmark \\
                     pe & \cellcolor{green!25}\cmark & \cellcolor{green!25}\cmark & \cellcolor{green!25}\cmark & \cellcolor{green!25}\cmark & \cellcolor{green!25}\cmark & \cellcolor{green!25}\cmark & \cellcolor{green!25}\cmark & \cellcolor{green!25}\cmark \\
       signal\_generator & \cellcolor{green!25}\cmark & \cellcolor{green!25}\cmark & \cellcolor{green!25}\cmark & \cellcolor{green!25}\cmark & \cellcolor{green!25}\cmark & \cellcolor{green!25}\cmark & \cellcolor{green!25}\cmark & \cellcolor{green!25}\cmark \\
            square\_wave & \cellcolor{green!25}\cmark & \cellcolor{green!25}\cmark & \cellcolor{green!25}\cmark & \cellcolor{green!25}\cmark & \cellcolor{green!25}\cmark & \cellcolor{green!25}\cmark & \cellcolor{green!25}\cmark & \cellcolor{green!25}\cmark \\
\hline
\multicolumn{1}{|c|}{\textbf{Syntax Success}} & \multicolumn{2}{c|}{\textbf{64\%}} & \multicolumn{2}{c|}{\textbf{68\%}} & \multicolumn{2}{c|}{\textbf{86\%}} & \multicolumn{2}{c|}{\textbf{78\%}} \\
\multicolumn{1}{|c|}{\textbf{Functionality Success}} & \multicolumn{2}{c|}{\textbf{40\%}} & \multicolumn{2}{c|}{\textbf{42\%}} & \multicolumn{2}{c|}{\textbf{60\%}} & \multicolumn{2}{c|}{\textbf{52\%}} \\
\hline
\end{tabular}
\end{table*}

\begin{table*}[!t]
\caption{Pass@1 Performance of HDLCoRe on RTLLM 2.0 Designs: Syntax and Functional Verification Results across Qwen2.5-Coder LLMs}
\label{tab:verification_results_c}
\centering
\begin{tabular}{|l|cc|cc|cc|}
\hline
\multirow{2}{*}{\textbf{Design}} & \multicolumn{2}{c|}{\textbf{Qwen2.5:7b-C}} & \multicolumn{2}{c|}{\textbf{Qwen2.5:14b-C}} & \multicolumn{2}{c|}{\textbf{Qwen2.5:32b-C}} \\
\cline{2-7}
 & Syn. & Fun. & Syn. & Fun. & Syn. & Fun. \\
\hline
                   accu & \cellcolor{green!25}\cmark &   \cellcolor{red!25}\xmark & \cellcolor{green!25}\cmark &   \cellcolor{red!25}\xmark & \cellcolor{green!25}\cmark & \cellcolor{green!25}\cmark \\
            adder\_8bit  & \cellcolor{green!25}\cmark & \cellcolor{green!25}\cmark &   \cellcolor{red!25}\xmark &   \cellcolor{red!25}\xmark & \cellcolor{green!25}\cmark & \cellcolor{green!25}\cmark \\
            adder\_16bit & \cellcolor{green!25}\cmark & \cellcolor{green!25}\cmark & \cellcolor{green!25}\cmark & \cellcolor{green!25}\cmark & \cellcolor{green!25}\cmark & \cellcolor{green!25}\cmark \\
            adder\_32bit & \cellcolor{green!25}\cmark &   \cellcolor{red!25}\xmark &   \cellcolor{red!25}\xmark &   \cellcolor{red!25}\xmark & \cellcolor{green!25}\cmark &   \cellcolor{red!25}\xmark \\
              adder\_bcd & \cellcolor{green!25}\cmark &   \cellcolor{red!25}\xmark & \cellcolor{green!25}\cmark & \cellcolor{green!25}\cmark & \cellcolor{green!25}\cmark & \cellcolor{green!25}\cmark \\
       adder\_pipe\_64bit & \cellcolor{green!25}\cmark &   \cellcolor{red!25}\xmark & \cellcolor{green!25}\cmark &   \cellcolor{red!25}\xmark & \cellcolor{green!25}\cmark & \cellcolor{green!25}\cmark \\
        comparator\_3bit & \cellcolor{green!25}\cmark & \cellcolor{green!25}\cmark & \cellcolor{green!25}\cmark & \cellcolor{green!25}\cmark & \cellcolor{green!25}\cmark & \cellcolor{green!25}\cmark \\
        comparator\_4bit & \cellcolor{green!25}\cmark &   \cellcolor{red!25}\xmark &   \cellcolor{red!25}\xmark &   \cellcolor{red!25}\xmark & \cellcolor{green!25}\cmark &   \cellcolor{red!25}\xmark \\
              div\_16bit & \cellcolor{green!25}\cmark &   \cellcolor{red!25}\xmark & \cellcolor{green!25}\cmark &   \cellcolor{red!25}\xmark & \cellcolor{green!25}\cmark &   \cellcolor{red!25}\xmark \\
             radix2\_div &   \cellcolor{red!25}\xmark &   \cellcolor{red!25}\xmark &   \cellcolor{red!25}\xmark &   \cellcolor{red!25}\xmark &   \cellcolor{red!25}\xmark &   \cellcolor{red!25}\xmark \\
             multi\_8bit & \cellcolor{green!25}\cmark & \cellcolor{green!25}\cmark & \cellcolor{green!25}\cmark &   \cellcolor{red!25}\xmark & \cellcolor{green!25}\cmark & \cellcolor{green!25}\cmark \\
            multi\_16bit & \cellcolor{green!25}\cmark &   \cellcolor{red!25}\xmark & \cellcolor{green!25}\cmark & \cellcolor{green!25}\cmark & \cellcolor{green!25}\cmark & \cellcolor{green!25}\cmark \\
       multi\_booth\_8bit & \cellcolor{green!25}\cmark & \cellcolor{green!25}\cmark & \cellcolor{green!25}\cmark & \cellcolor{green!25}\cmark & \cellcolor{green!25}\cmark & \cellcolor{green!25}\cmark \\
        multi\_pipe\_4bit & \cellcolor{green!25}\cmark &   \cellcolor{red!25}\xmark & \cellcolor{green!25}\cmark &   \cellcolor{red!25}\xmark & \cellcolor{green!25}\cmark &   \cellcolor{red!25}\xmark \\
        multi\_pipe\_8bit &   \cellcolor{red!25}\xmark &   \cellcolor{red!25}\xmark &   \cellcolor{red!25}\xmark &   \cellcolor{red!25}\xmark &   \cellcolor{red!25}\xmark &   \cellcolor{red!25}\xmark \\
      fixed\_point\_adder & \cellcolor{green!25}\cmark & \cellcolor{green!25}\cmark & \cellcolor{green!25}\cmark & \cellcolor{green!25}\cmark & \cellcolor{green!25}\cmark & \cellcolor{green!25}\cmark \\
fixed\_point\_substractor & \cellcolor{green!25}\cmark & \cellcolor{green!25}\cmark & \cellcolor{green!25}\cmark & \cellcolor{green!25}\cmark & \cellcolor{green!25}\cmark & \cellcolor{green!25}\cmark \\
            float\_multi & \cellcolor{green!25}\cmark &   \cellcolor{red!25}\xmark &   \cellcolor{red!25}\xmark &   \cellcolor{red!25}\xmark &   \cellcolor{red!25}\xmark &   \cellcolor{red!25}\xmark \\
              sub\_64bit & \cellcolor{green!25}\cmark & \cellcolor{green!25}\cmark & \cellcolor{green!25}\cmark & \cellcolor{green!25}\cmark & \cellcolor{green!25}\cmark & \cellcolor{green!25}\cmark \\
\hline
             counter\_12 & \cellcolor{green!25}\cmark & \cellcolor{green!25}\cmark & \cellcolor{green!25}\cmark & \cellcolor{green!25}\cmark & \cellcolor{green!25}\cmark & \cellcolor{green!25}\cmark \\
             JC\_counter & \cellcolor{green!25}\cmark & \cellcolor{green!25}\cmark & \cellcolor{green!25}\cmark & \cellcolor{green!25}\cmark & \cellcolor{green!25}\cmark & \cellcolor{green!25}\cmark \\
           ring\_counter & \cellcolor{green!25}\cmark & \cellcolor{green!25}\cmark & \cellcolor{green!25}\cmark & \cellcolor{green!25}\cmark & \cellcolor{green!25}\cmark & \cellcolor{green!25}\cmark \\
       up\_down\_counter  & \cellcolor{green!25}\cmark & \cellcolor{green!25}\cmark & \cellcolor{green!25}\cmark & \cellcolor{green!25}\cmark & \cellcolor{green!25}\cmark & \cellcolor{green!25}\cmark \\
                    fsm & \cellcolor{green!25}\cmark & \cellcolor{green!25}\cmark &   \cellcolor{red!25}\xmark &   \cellcolor{red!25}\xmark & \cellcolor{green!25}\cmark &   \cellcolor{red!25}\xmark \\
      sequence\_detector &   \cellcolor{red!25}\xmark &   \cellcolor{red!25}\xmark &   \cellcolor{red!25}\xmark &   \cellcolor{red!25}\xmark &   \cellcolor{red!25}\xmark &   \cellcolor{red!25}\xmark \\
          right\_shifter & \cellcolor{green!25}\cmark &   \cellcolor{red!25}\xmark & \cellcolor{green!25}\cmark & \cellcolor{green!25}\cmark & \cellcolor{green!25}\cmark & \cellcolor{green!25}\cmark \\
               freq\_div &   \cellcolor{red!25}\xmark &   \cellcolor{red!25}\xmark & \cellcolor{green!25}\cmark & \cellcolor{green!25}\cmark & \cellcolor{green!25}\cmark &   \cellcolor{red!25}\xmark \\
         freq\_divbyeven &   \cellcolor{red!25}\xmark &   \cellcolor{red!25}\xmark &   \cellcolor{red!25}\xmark &   \cellcolor{red!25}\xmark &   \cellcolor{red!25}\xmark &   \cellcolor{red!25}\xmark \\
         freq\_divbyfrac & \cellcolor{green!25}\cmark &   \cellcolor{red!25}\xmark & \cellcolor{green!25}\cmark &   \cellcolor{red!25}\xmark & \cellcolor{green!25}\cmark &   \cellcolor{red!25}\xmark \\
          freq\_divbyodd &   \cellcolor{red!25}\xmark &   \cellcolor{red!25}\xmark & \cellcolor{green!25}\cmark &   \cellcolor{red!25}\xmark & \cellcolor{green!25}\cmark &   \cellcolor{red!25}\xmark \\
\hline
              asyn\_fifo &   \cellcolor{red!25}\xmark &   \cellcolor{red!25}\xmark &   \cellcolor{red!25}\xmark &   \cellcolor{red!25}\xmark &   \cellcolor{red!25}\xmark &   \cellcolor{red!25}\xmark \\
             LIFObuffer &   \cellcolor{red!25}\xmark &   \cellcolor{red!25}\xmark &   \cellcolor{red!25}\xmark &   \cellcolor{red!25}\xmark & \cellcolor{green!25}\cmark &   \cellcolor{red!25}\xmark \\
         barrel\_shifter & \cellcolor{green!25}\cmark &   \cellcolor{red!25}\xmark & \cellcolor{green!25}\cmark &   \cellcolor{red!25}\xmark & \cellcolor{green!25}\cmark &   \cellcolor{red!25}\xmark \\
                   LFSR &   \cellcolor{red!25}\xmark &   \cellcolor{red!25}\xmark &   \cellcolor{red!25}\xmark &   \cellcolor{red!25}\xmark &   \cellcolor{red!25}\xmark &   \cellcolor{red!25}\xmark \\
                    RAM &   \cellcolor{red!25}\xmark &   \cellcolor{red!25}\xmark &   \cellcolor{red!25}\xmark &   \cellcolor{red!25}\xmark & \cellcolor{green!25}\cmark & \cellcolor{green!25}\cmark \\
                    ROM & \cellcolor{green!25}\cmark & \cellcolor{green!25}\cmark & \cellcolor{green!25}\cmark & \cellcolor{green!25}\cmark & \cellcolor{green!25}\cmark & \cellcolor{green!25}\cmark \\
\hline
               calendar & \cellcolor{green!25}\cmark & \cellcolor{green!25}\cmark & \cellcolor{green!25}\cmark & \cellcolor{green!25}\cmark & \cellcolor{green!25}\cmark & \cellcolor{green!25}\cmark \\
            edge\_detect & \cellcolor{green!25}\cmark &   \cellcolor{red!25}\xmark & \cellcolor{green!25}\cmark & \cellcolor{green!25}\cmark & \cellcolor{green!25}\cmark & \cellcolor{green!25}\cmark \\
        parallel2serial & \cellcolor{green!25}\cmark &   \cellcolor{red!25}\xmark & \cellcolor{green!25}\cmark & \cellcolor{green!25}\cmark & \cellcolor{green!25}\cmark &   \cellcolor{red!25}\xmark \\
           pulse\_detect & \cellcolor{green!25}\cmark &   \cellcolor{red!25}\xmark & \cellcolor{green!25}\cmark &   \cellcolor{red!25}\xmark & \cellcolor{green!25}\cmark &   \cellcolor{red!25}\xmark \\
        serial2parallel & \cellcolor{green!25}\cmark &   \cellcolor{red!25}\xmark & \cellcolor{green!25}\cmark &   \cellcolor{red!25}\xmark & \cellcolor{green!25}\cmark &   \cellcolor{red!25}\xmark \\
           synchronizer & \cellcolor{green!25}\cmark & \cellcolor{green!25}\cmark & \cellcolor{green!25}\cmark & \cellcolor{green!25}\cmark & \cellcolor{green!25}\cmark & \cellcolor{green!25}\cmark \\
          traffic\_light & \cellcolor{green!25}\cmark &   \cellcolor{red!25}\xmark & \cellcolor{green!25}\cmark &   \cellcolor{red!25}\xmark & \cellcolor{green!25}\cmark &   \cellcolor{red!25}\xmark \\
            width\_8to16 &   \cellcolor{red!25}\xmark &   \cellcolor{red!25}\xmark & \cellcolor{green!25}\cmark & \cellcolor{green!25}\cmark & \cellcolor{green!25}\cmark & \cellcolor{green!25}\cmark \\
                    alu &   \cellcolor{red!25}\xmark &   \cellcolor{red!25}\xmark & \cellcolor{green!25}\cmark &   \cellcolor{red!25}\xmark &   \cellcolor{red!25}\xmark &   \cellcolor{red!25}\xmark \\
           clkgenerator &   \cellcolor{red!25}\xmark &   \cellcolor{red!25}\xmark & \cellcolor{green!25}\cmark &   \cellcolor{red!25}\xmark & \cellcolor{green!25}\cmark & \cellcolor{green!25}\cmark \\
              instr\_reg & \cellcolor{green!25}\cmark & \cellcolor{green!25}\cmark & \cellcolor{green!25}\cmark & \cellcolor{green!25}\cmark & \cellcolor{green!25}\cmark & \cellcolor{green!25}\cmark \\
                     pe & \cellcolor{green!25}\cmark & \cellcolor{green!25}\cmark & \cellcolor{green!25}\cmark & \cellcolor{green!25}\cmark & \cellcolor{green!25}\cmark & \cellcolor{green!25}\cmark \\
       signal\_generator & \cellcolor{green!25}\cmark & \cellcolor{green!25}\cmark & \cellcolor{green!25}\cmark & \cellcolor{green!25}\cmark & \cellcolor{green!25}\cmark &   \cellcolor{red!25}\xmark \\
            square\_wave & \cellcolor{green!25}\cmark & \cellcolor{green!25}\cmark & \cellcolor{green!25}\cmark & \cellcolor{green!25}\cmark & \cellcolor{green!25}\cmark & \cellcolor{green!25}\cmark \\
\hline
\multicolumn{1}{|c|}{\textbf{Syntax Success}} & \multicolumn{2}{c|}{\textbf{74\%}} & \multicolumn{2}{c|}{\textbf{74\%}} & \multicolumn{2}{c|}{\textbf{84\%}} \\
\multicolumn{1}{|c|}{\textbf{Functionality Success}} & \multicolumn{2}{c|}{\textbf{40\%}} & \multicolumn{2}{c|}{\textbf{48\%}} & \multicolumn{2}{c|}{\textbf{54\%}} \\
\hline
\end{tabular}
\end{table*}

\begin{itemize}
    \item \textbf{Lines of Code (LoC):} The number of lines in the HDL code (Verilog) of the reference design. This metric reflects the implementation complexity from a human designer's perspective. Designs with higher LoC values typically require more effort to implement correctly and are more challenging for LLMs to generate.
    
    \item \textbf{Cells:} The number of logic cells in the synthesized gate-level netlist. This metric represents the actual hardware complexity and scale of the design. A higher cell count indicates a larger design in terms of area and power consumption. The cell counts range from as few as 6 to 2,435, demonstrating the wide spectrum of design scales covered in this benchmark.
\end{itemize}

As shown in the table, RTLLM 2.0 covers a diverse range of designs across arithmetic operations, memory elements, control logic, and miscellaneous functional units. The benchmark includes both relatively simple components (e.g., an 8-bit adder with 26 LoC) and more complex designs (e.g., a 64-bit pipeline adder with 197 LoC), enabling comprehensive evaluation of LLMs' capabilities across various design complexity levels.

\begin{figure*}[!t]
    \centering
    \includegraphics[width=\textwidth]{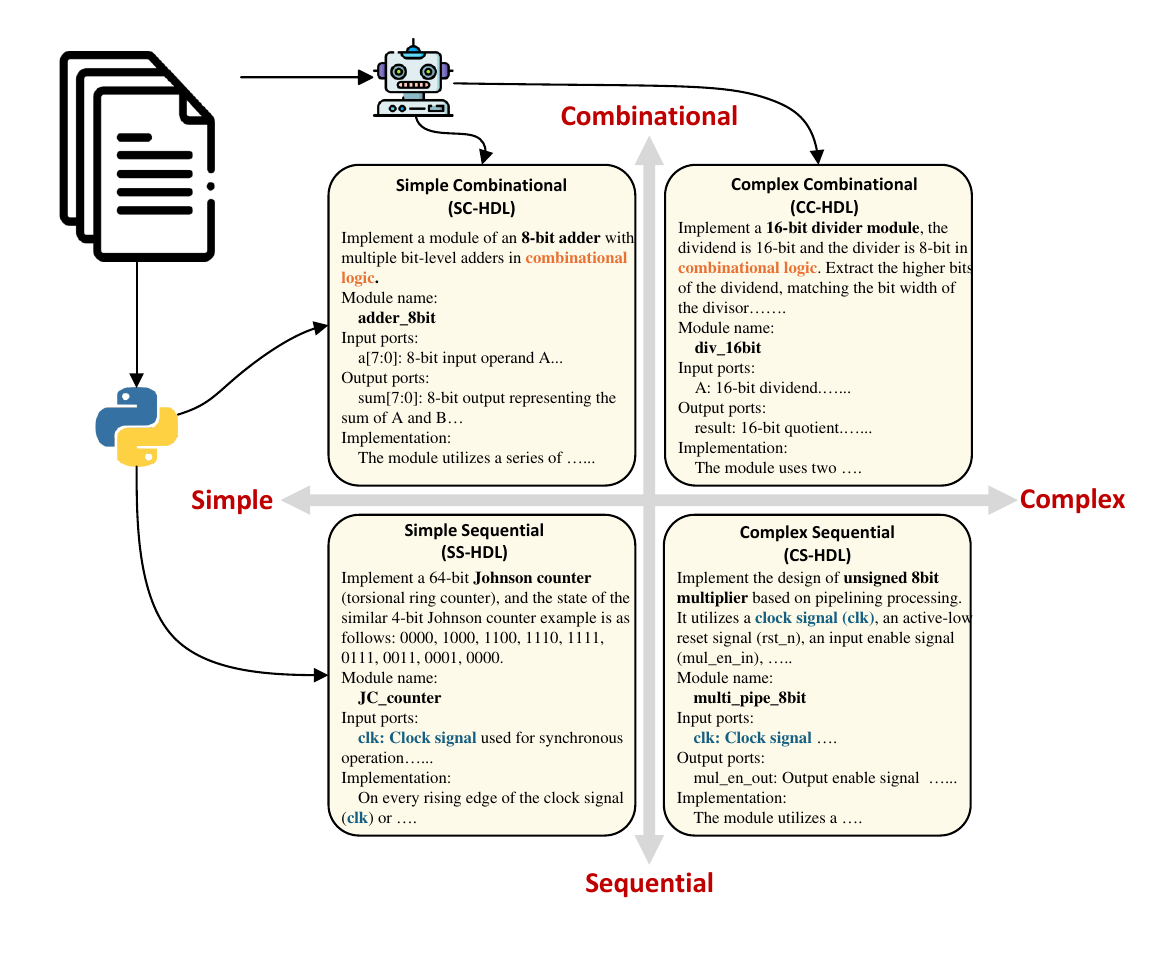}
    \caption{\textbf{CoT Classification mechanism}. We develop scripts that automatically classify \textbf{the logical category} of each problem description. Meanwhile, the LLM autonomously evaluates the \textbf{complexity} of each problem based on its internal knowledge.}
    \label{fig:cot-classification}
\end{figure*}

\section{Experiment Result Supplement}

As shown in Table~\ref{tab:verification_results}, our HDLCoRe framework demonstrates a clear pattern of success across the RTLLM 2.0 benchmark designs. The results reveal that simpler designs such as calendar, square\_wave, and basic adders are consistently generated correctly across most model variants, achieving both syntax and functional correctness. In contrast, more complex designs like asyn\_fifo, sequence\_detector, and multi\_pipe\_8bit prove challenging for all models, with few or no models able to generate functionally correct implementations. This pattern indicates that design complexity significantly impacts generation success, with HDL code for designs having lower Lines of Code (LoC) and cell counts generally being more successfully generated.

\section{Specify CoT based on Task Type and Complexity Supplement}
For each problem, our framework initially determines the appropriate CoT strategies by considering both its logical category and complexity. As illustrated in Figure~\ref{fig:cot-classification}, we perform a two-step classification process on each problem. In the first step, we designed a script to extract relevant keywords (e.g., “combinational logic” or “clk”) from the description, effectively categorizing the problem as either combinational or sequential logic. Since different LLMs have varying capabilities in RTL generation, we leverage the problem’s assessed complexity to tailor the CoT strategies. We pass the problem description to the LLM, allowing it to self-assess the difficulty of the problem. By combining the logic category with the LLM’s complexity evaluation, we classify problems into four types: Simple-Combinational (SC-HDL), Complex-Combinational (CC-HDL), Simple-Sequential (SS-HDL), and Complex-Sequential (CS-HDL), each corresponding to a specific CoT strategy for RTL generation.

\end{document}